\documentclass[10pt,twocolumn,letterpaper]{article}

\usepackage{cvpr}
\usepackage{times}
\usepackage{epsfig}
\usepackage{graphicx}
\usepackage{amsmath}
\usepackage{amssymb}

\usepackage{caption}
\usepackage{subcaption}

\usepackage{booktabs} %
\usepackage{tabularx} %
\usepackage{url}

 \makeatletter
 \DeclareRobustCommand\onedot{\futurelet\@let@token\@onedot}
 \def\@onedot{\ifx\@let@token.\else.\null\fi\xspace}
 \def\eg{e.g\onedot} \def\Eg{E.g\onedot}
 \def\ie{i.e\onedot}

 \makeatother

\DeclareRobustCommand{\Figref}[1]{Figure~\ref{#1}}

\DeclareRobustCommand{\secref}[1]{Section~\ref{#1}}

\newcommand{\nMoviesAD}{46\xspace}
\newcommand{\nMoviesScript}{26\xspace}

\usepackage[pagebackref=true,breaklinks=true,letterpaper=true,colorlinks,bookmarks=false]{hyperref}

 \makeatletter
 \DeclareRobustCommand\onedot{\futurelet\@let@token\@onedot}
 \def\@onedot{\ifx\@let@token.\else.\null\fi\xspace}
 \def\eg{e.g\onedot} \def\Eg{E.g\onedot}
 \def\ie{i.e\onedot}

 \makeatother

\DeclareRobustCommand{\Figref}[1]{Figure~\ref{#1}}

\DeclareRobustCommand{\secref}[1]{Section~\ref{#1}}

\newcommand{\invisible}[1]{}%

\newcommand{\figvspace}{\vspace{-.5cm}}

\graphicspath{{./fig/}{./fig/plots/}}
\newcommand{\redtext}[1]{\emph{\textcolor{red}{#1}}}

\cvprfinalcopy %

\ifcvprfinal\fi
\begin{document}

\title{A Dataset for Movie Description}
\newcommand{\authSpace}{&}
\author{\begin{tabular}{cccc}
Anna Rohrbach$^{1}$ \authSpace Marcus Rohrbach$^{2}$ \authSpace Niket Tandon$^{1}$ \authSpace Bernt Schiele$^{1}$\\
\end{tabular}\\
\begin{tabular}{cccc}
\multicolumn{4}{c}{$^{1}$Max Planck Institute for Informatics, Saarbr{\"u}cken, Germany}\\
\multicolumn{4}{c}{$^{2}$UC Berkeley EECS and ICSI, Berkeley, CA, United States}\\
\end{tabular}}

\maketitle
\thispagestyle{plain}
\pagestyle{plain}

\begin{abstract}
Descriptive video service (DVS) provides linguistic descriptions of movies and allows visually impaired people to follow a movie along with their peers. Such descriptions are by design mainly visual and thus naturally form an interesting data source for computer vision and computational linguistics. In this work we propose a novel dataset which contains transcribed DVS, which is temporally aligned to full length HD movies. In addition we also collected the aligned movie scripts which have been used in prior work and compare the two different sources of descriptions. In total the \emph{Movie Description} dataset contains a parallel corpus of over 54,000 sentences and video snippets from 72 HD movies. We characterize the dataset by benchmarking different approaches for generating video descriptions. Comparing DVS to scripts, we find that DVS is far more visual and describes precisely what \emph{is shown} rather than what \emph{should happen} according to the scripts created prior to movie production.
\end{abstract}

\section{Introduction\invisible{ - 1.5 pages}}

\begin{figure}[t]
\scriptsize
\begin{center}
\begin{tabular}{@{}p{2.5cm}p{2.5cm}p{2.5cm}}
\includegraphics[width=\linewidth]{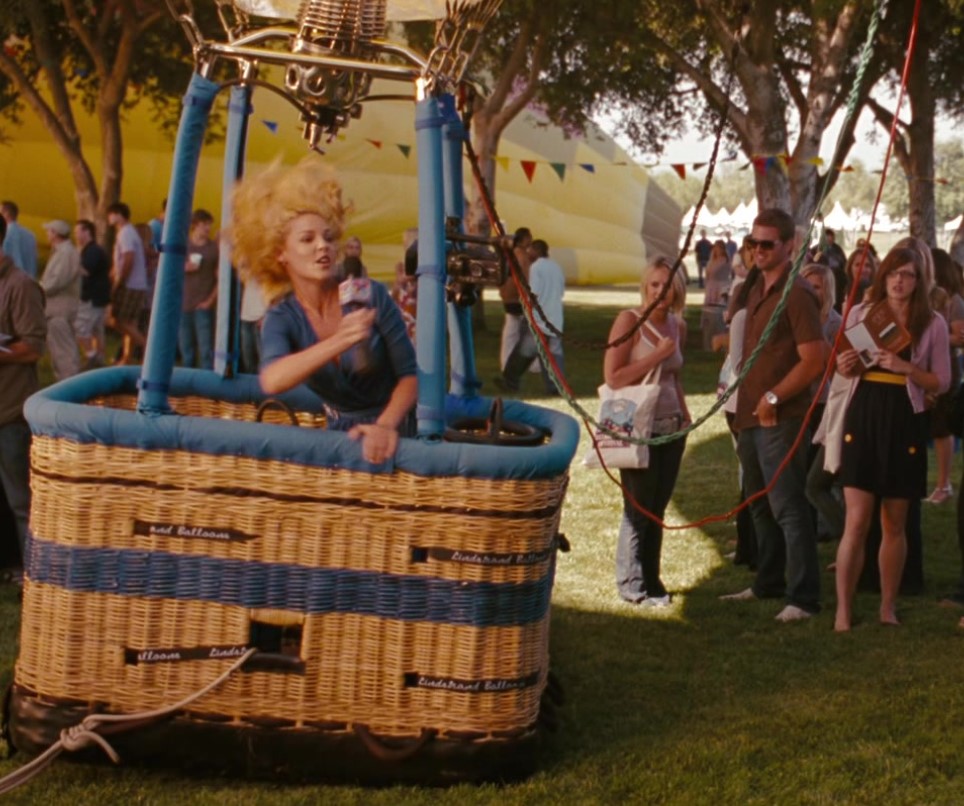} & \includegraphics[width=\linewidth]{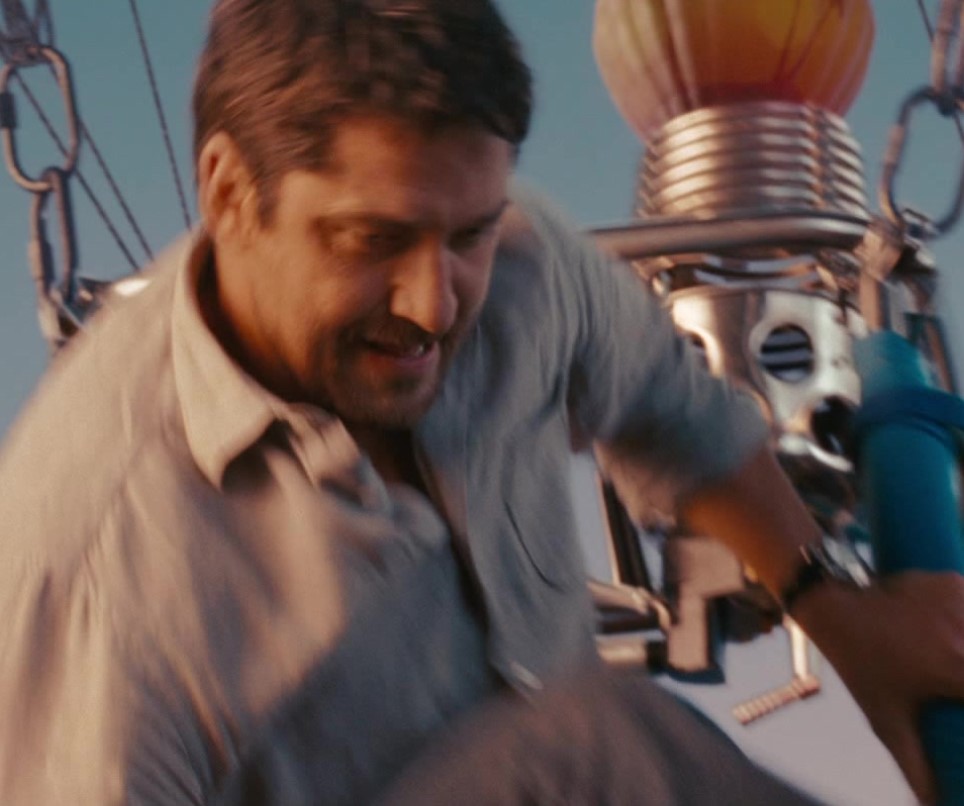} & \includegraphics[width=\linewidth]{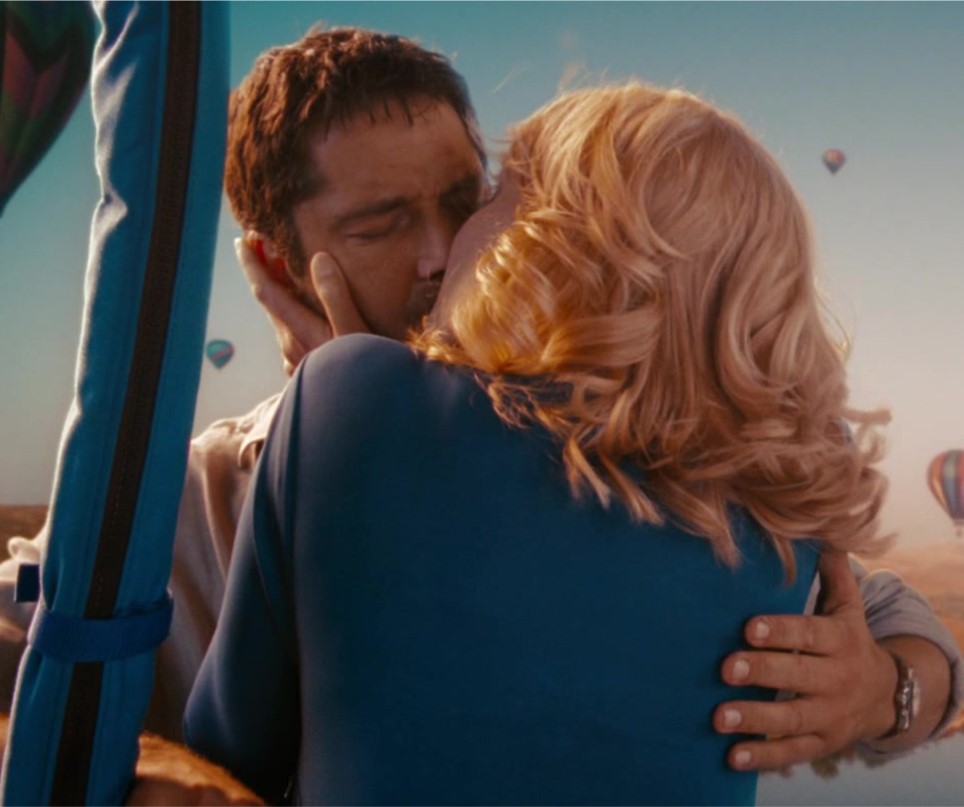} \\
\textbf{DVS}: Abby gets in the basket. & Mike leans over and sees how high they are. & Abby clasps her hands around his face and kisses him passionately. \\
\textbf{Script}: After a moment a frazzled Abby pops up in his place. & Mike looks down to see -- they are now fifteen feet above the ground. & For the first time in her life, she stops thinking and grabs Mike and kisses the hell out of him. \\
\end{tabular}
\caption{Audio descriptions (DVS - descriptive video service),  movie scripts (scripts) from the movie ``Ugly Truth''.}
\label{fig:teaser1}
\end{center}
\end{figure}

Audio descriptions (DVS - descriptive video service) make movies accessible to millions of blind or visually impaired people\footnote{\label{fn:blind} In this work we refer for simplicity to ``the blind'' to account for all blind and visually impaired people which benefit from DVS, knowing of the variety of visually impaired and that DVS is not accessible to all.}. DVS provides an audio narrative of the ``most important aspects of the visual information'' \cite{salway07corpus}, namely actions, gestures, scenes, and character appearance as can be seen in Figures \ref{fig:teaser1} and \ref{fig:teaser}. DVS is prepared by trained describers and read by professional narrators. More and more movies are audio transcribed, but it may take up to 60 person-hours to describe a 2-hour movie \cite{lakritz06tr}, resulting in the fact that only a small subset of movies and TV programs are available for the blind. Consequently, automating this would be a noble task.

In addition to the benefits for the blind, generating descriptions for video is an interesting task in itself requiring to understand and combine core techniques of computer vision and computational linguistics. To understand the visual input one has to reliably recognize scenes, human activities, and participating objects. To generate a good description one has to decide what part of the visual information to verbalize, \ie recognize what is salient.
 
Large datasets of objects \cite{deng09cvpr} and scenes \cite{xiao10cvpr,zhou14nips}  had an important impact in the field and significantly improved our ability to recognize objects and scenes in combination with CNNs \cite{krizhevsky12nips}.
To be able to learn how to generate descriptions of visual content, parallel datasets of visual content paired with descriptions are indispensable~\cite{rohrbach13iccv}. While recently several large datasets have been released which provide images with descriptions \cite{ordonez11nips,flickr30k,coco2014}, video description datasets focus on short video snippets only and are limited in size \cite{chen11acl} or not publicly available \cite{over12tv}.
TACoS Multi-Level \cite{rohrbach14gcpr} and YouCook \cite{das13cvpr} are exceptions by providing multiple sentence descriptions and longer videos, however they are restricted to the cooking scenario.
In contrast, the data available with DVS provides realistic, open domain video paired with multiple sentence descriptions. It even goes beyond this by telling a story which means it allows to study how to extract plots and understand long term semantic dependencies and human interactions from the visual and textual data. 

\renewcommand{\bottomfraction}{0.8}
\setcounter{dbltopnumber}{2}
\renewcommand{\textfraction}{0.07}
\newcommand{\colwidth}{3.1cm}
\begin{figure*}[t]
\scriptsize
\begin{center}
\begin{tabular}{@{}p{\colwidth}p{\colwidth}p{\colwidth}p{\colwidth}p{\colwidth}}
 \includegraphics[width=\linewidth]{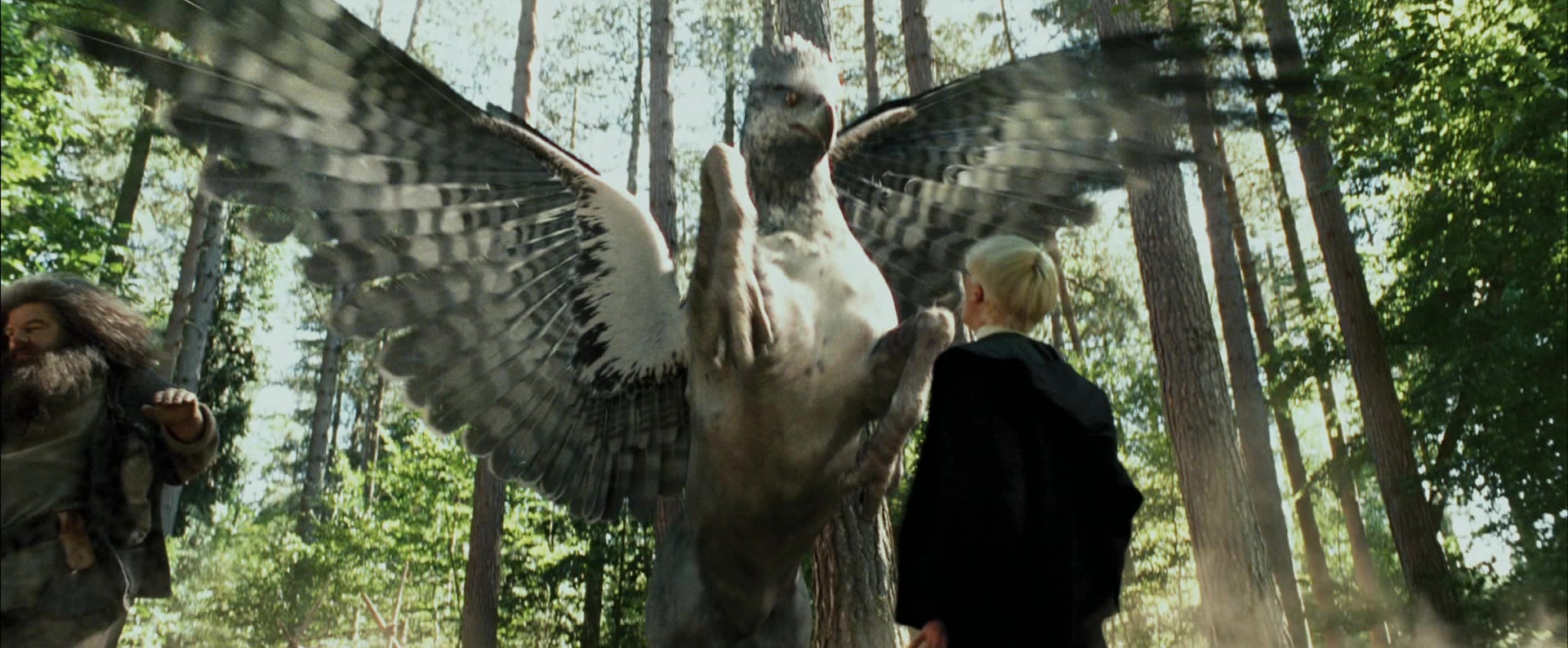} & \includegraphics[width=\linewidth]{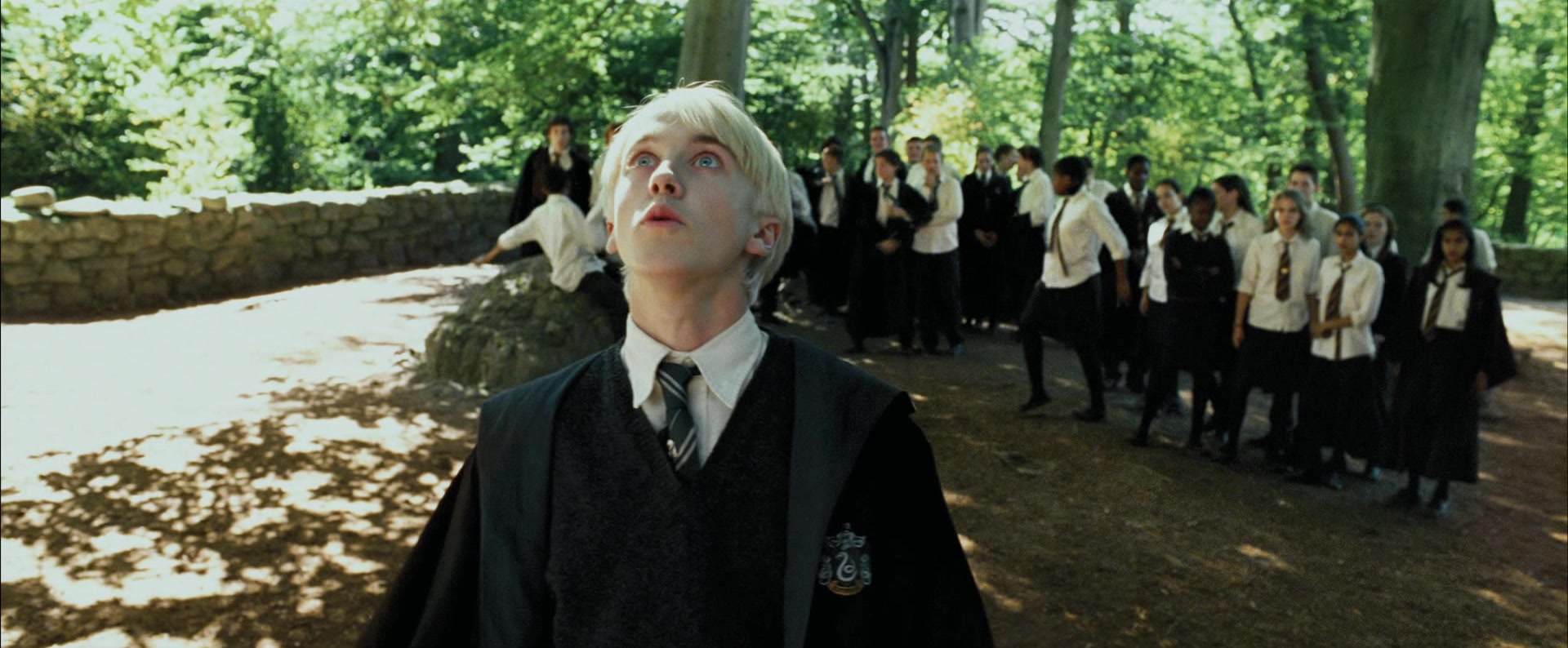} & \includegraphics[width=\linewidth]{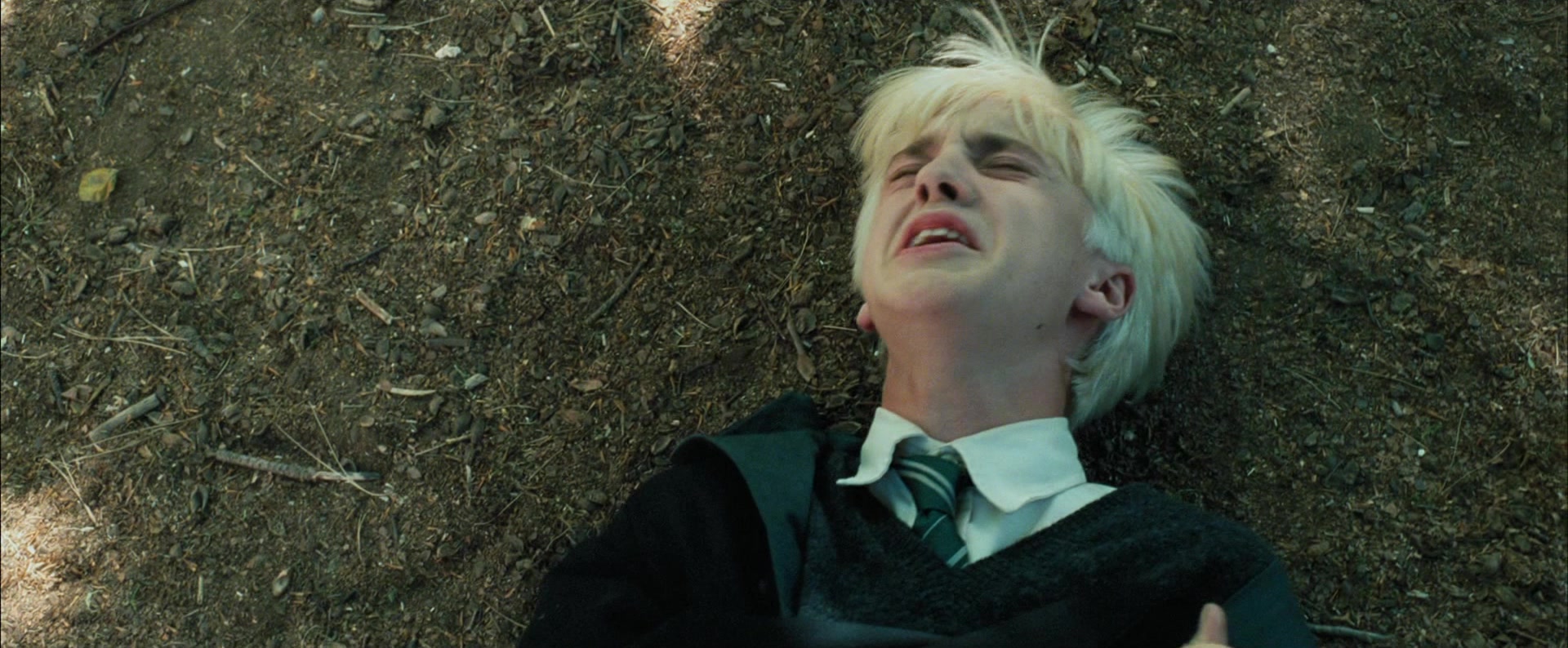} & \includegraphics[width=\linewidth]{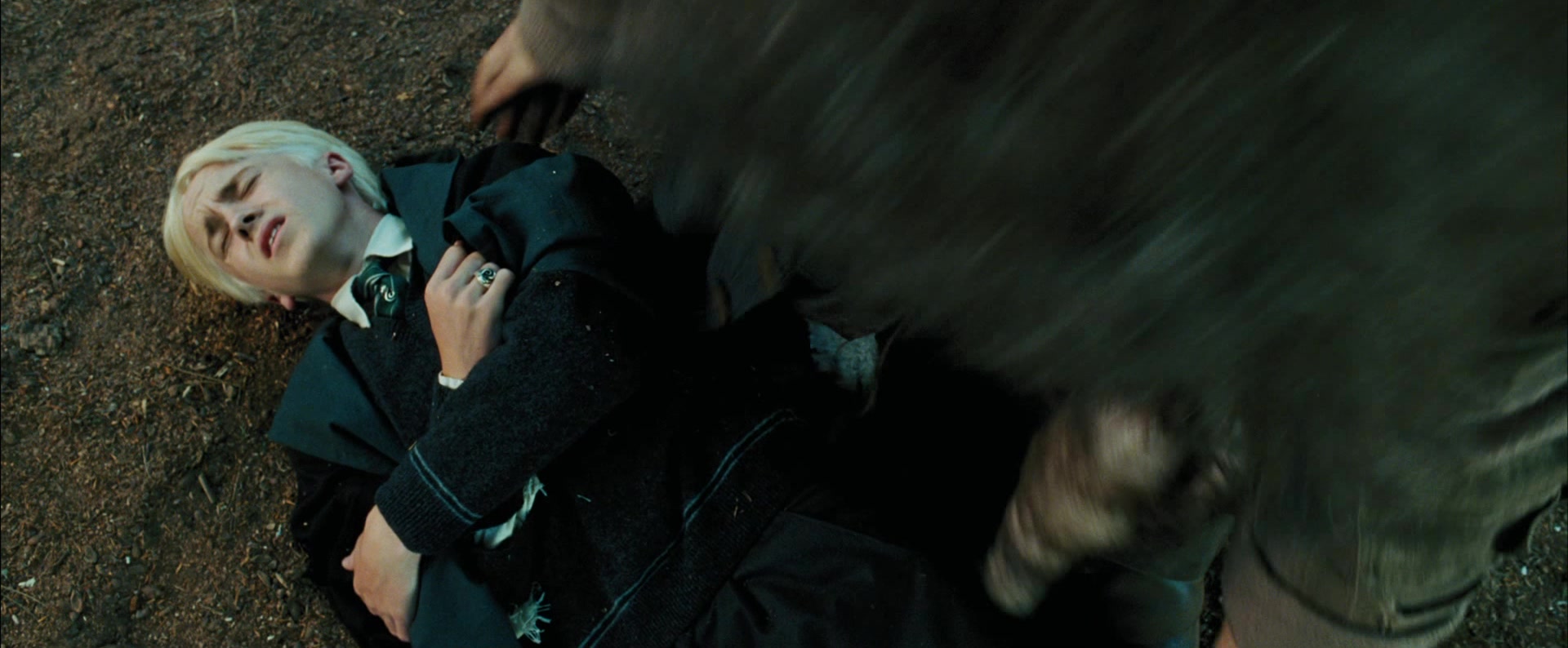} & \includegraphics[width=\linewidth]{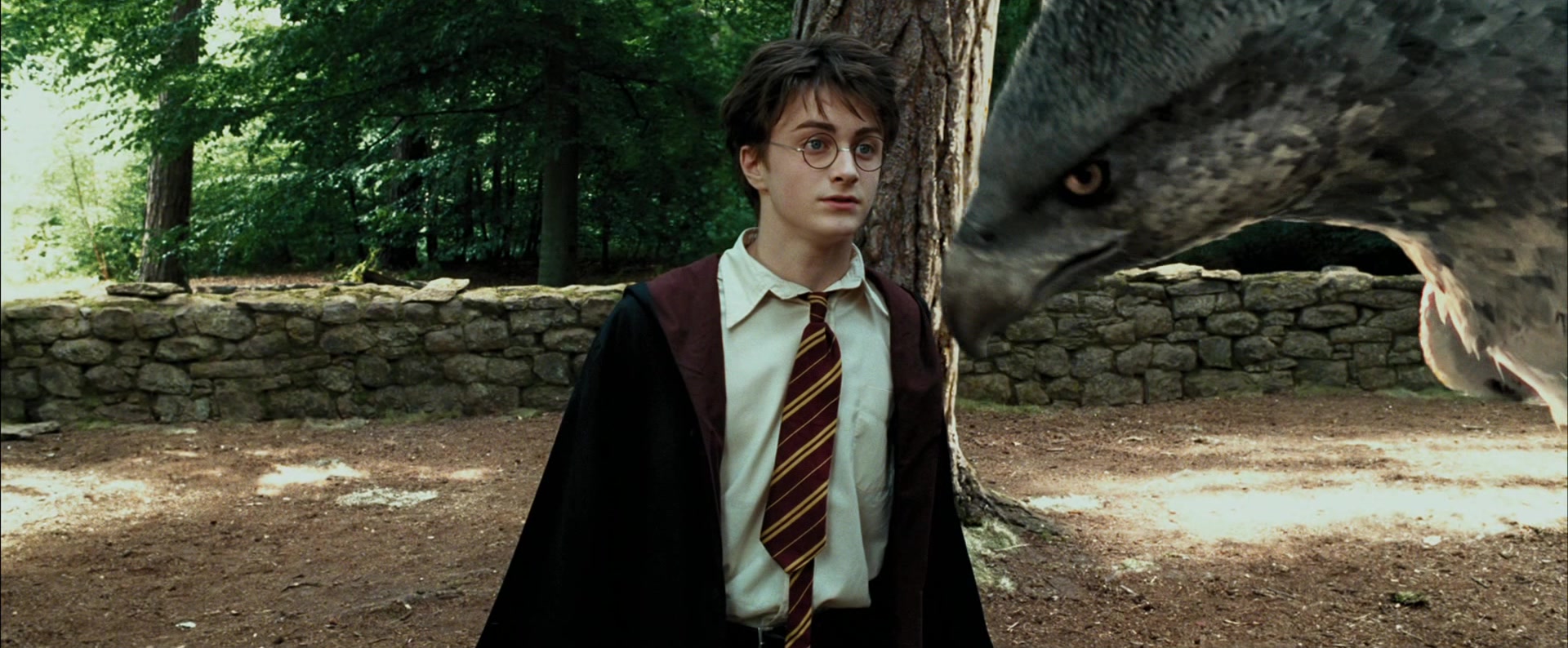}\\
 \textbf{DVS}: Buckbeak rears and attacks Malfoy. &  & & Hagrid lifts Malfoy up. & As Hagrid carries Malfoy away, the hippogriff gently nudges Harry. \\
 \textbf{Script}: In a flash, Buckbeak's steely talons slash down. & Malfoy freezes. & \redtext{Looks down at the blood blossoming on his robes.} & & \redtext{Buckbeak whips around, raises its talons and - seeing Harry - lowers them.}\\

 \includegraphics[width=\linewidth]{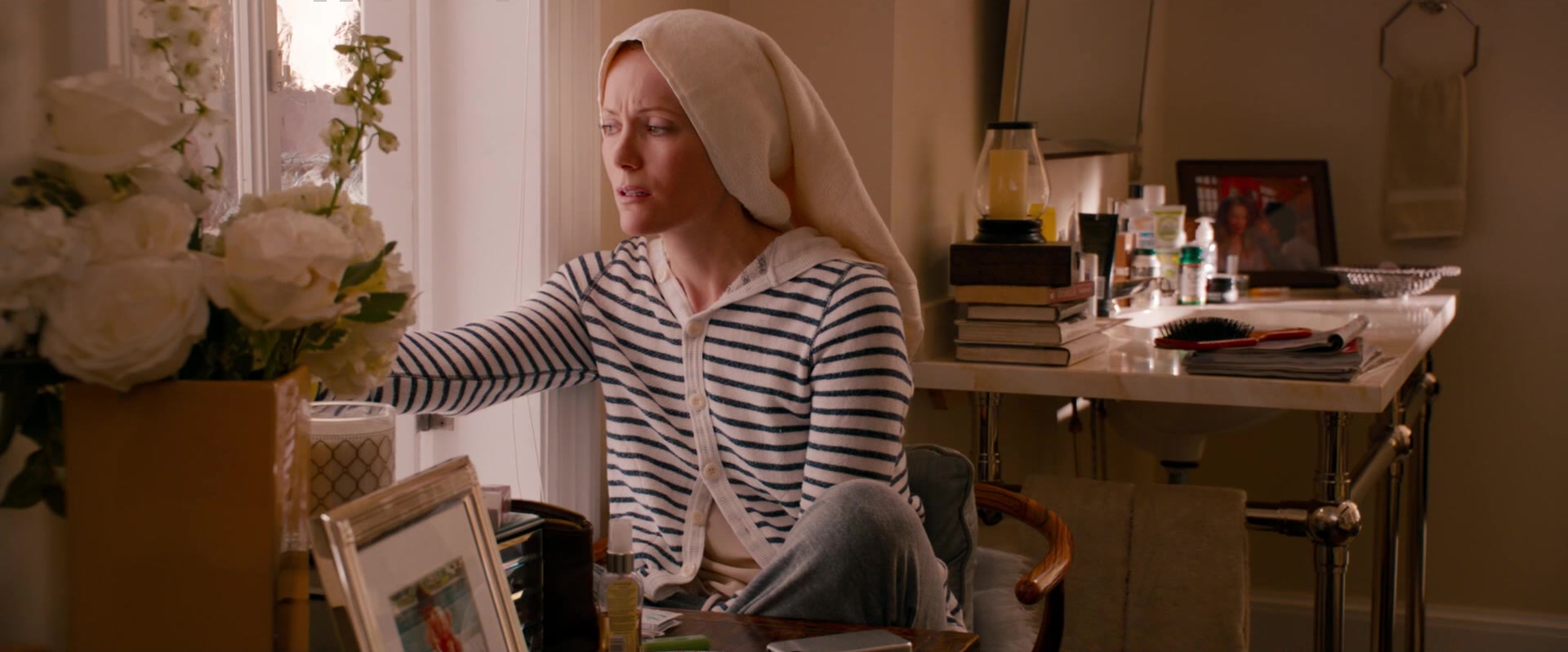} & \includegraphics[width=\linewidth]{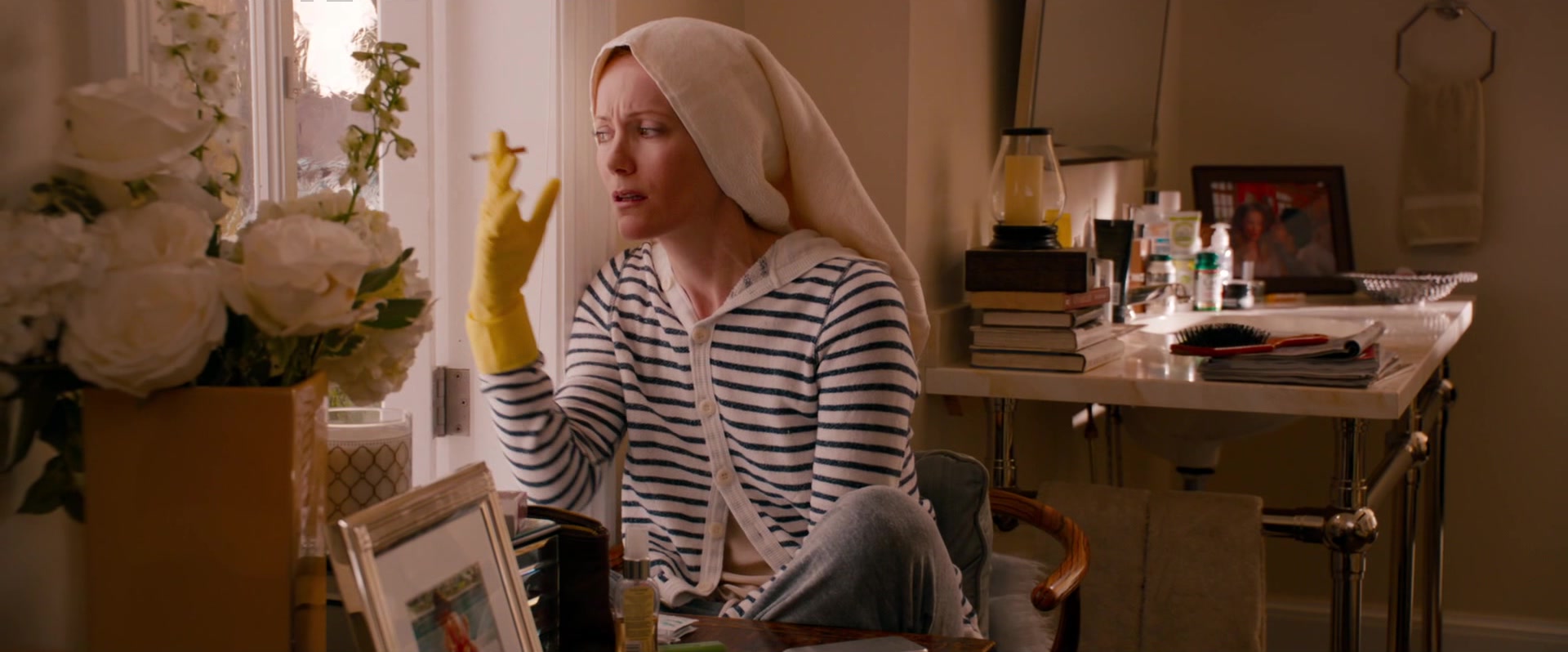} & \includegraphics[width=\linewidth]{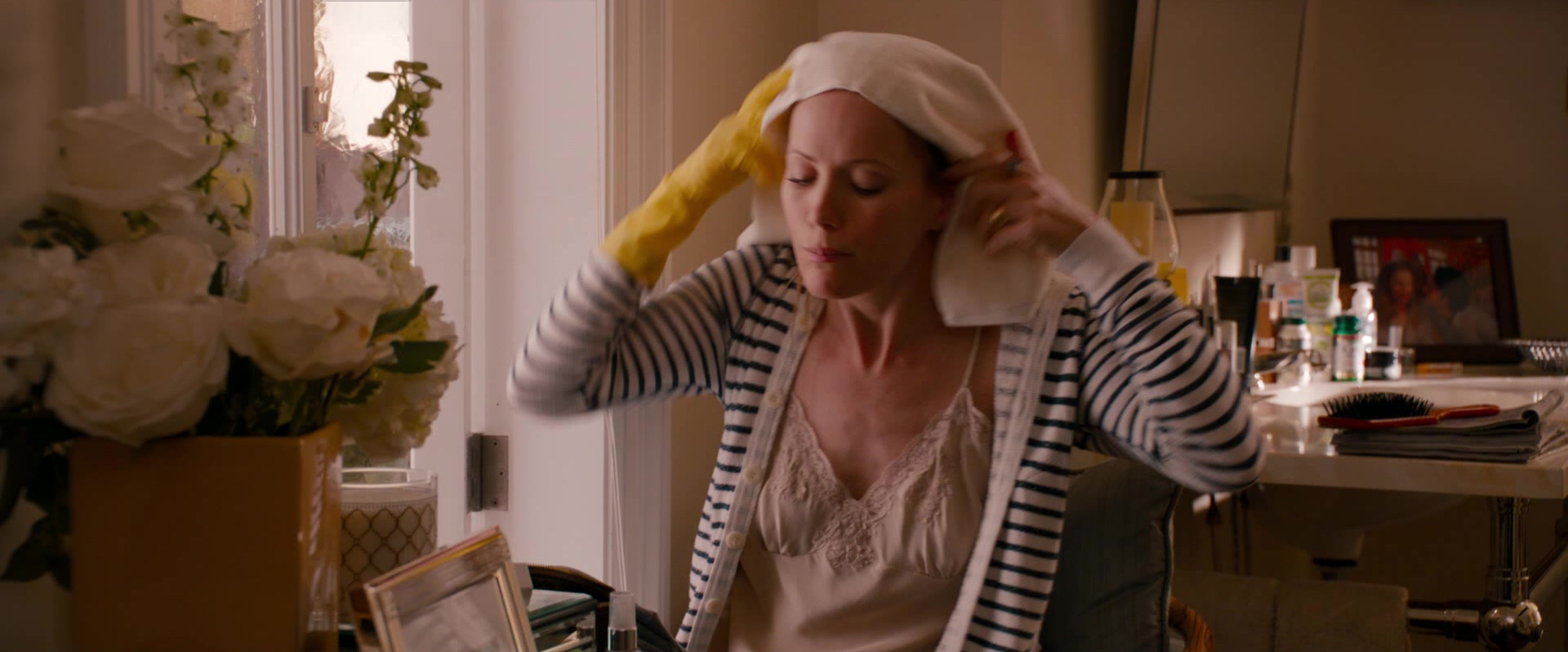} & \includegraphics[width=\linewidth]{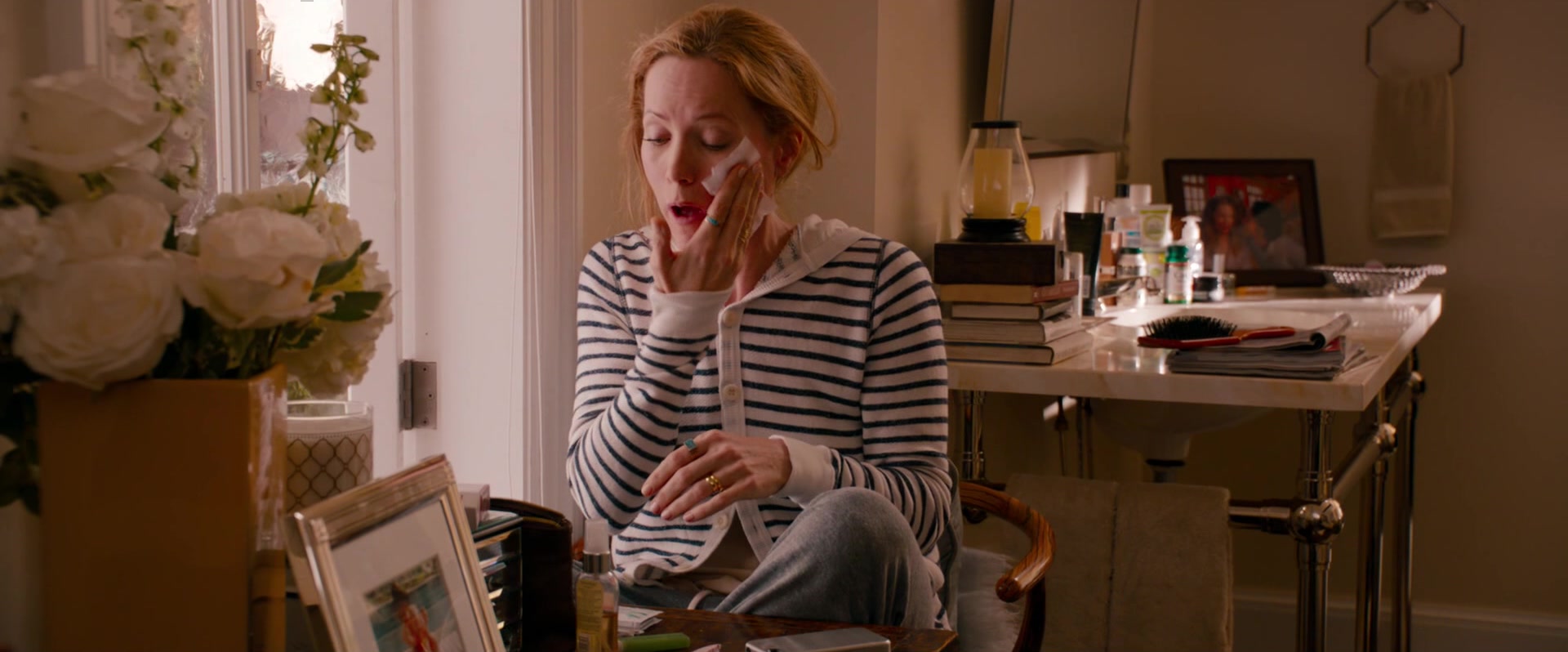} & \includegraphics[width=\linewidth]{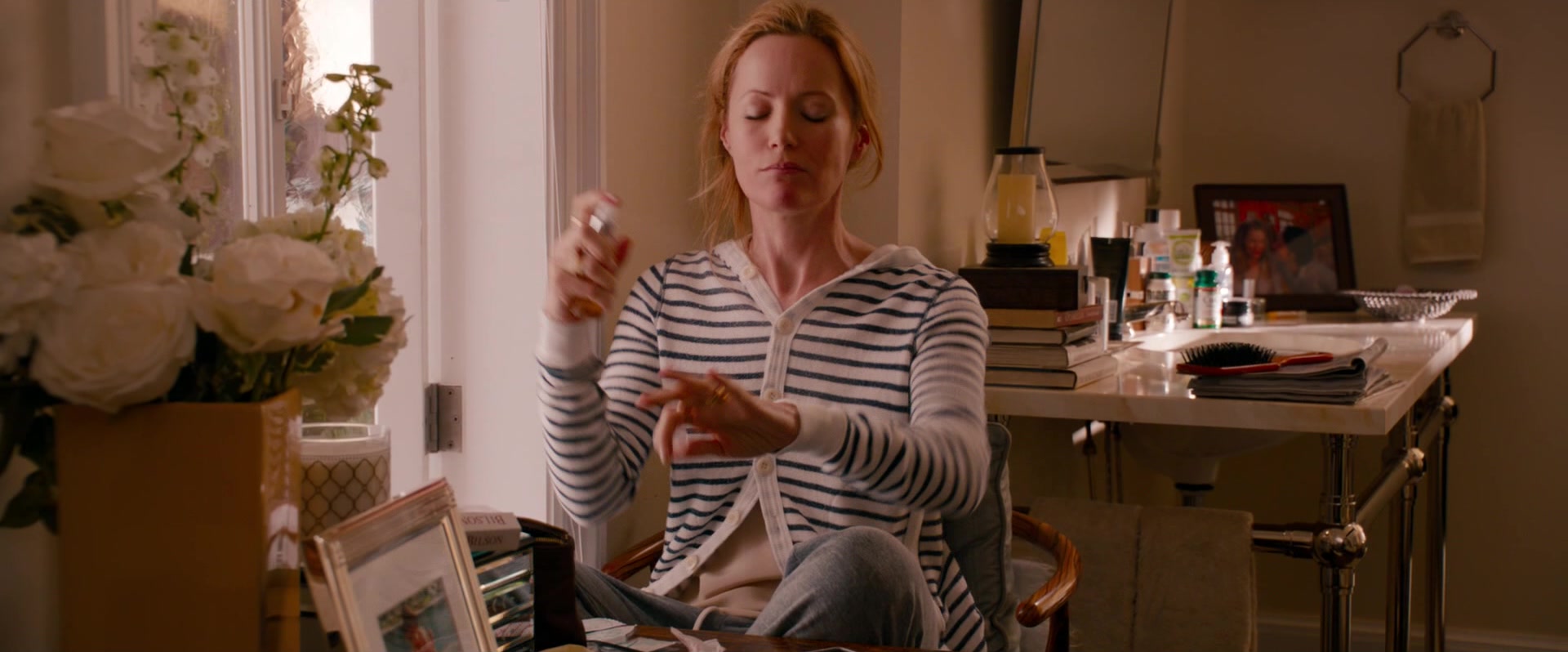}\\
 \textbf{DVS}: Another room, the wife and mother sits at a window with a towel over her hair. & She smokes a cigarette with a latex-gloved hand. & Putting the cigarette out, she uncovers her hair, removes the glove and pops gum in her mouth. & She pats her face and hands with a wipe, then sprays herself with perfume. & She pats her face and hands with a wipe, then sprays herself with perfume. \\
 \textbf{Script}: Debbie opens a window and sneaks a cigarette. & She holds her cigarette with a yellow dish washing glove. &  She puts out the cigarette and goes through an elaborate routine of hiding the smell of smoke. & She \redtext{puts some weird oil in her hair} and uses a wet nap on her neck \redtext{and clothes and brushes her teeth}. & She sprays cologne \redtext{and walks through it.}\\

 \includegraphics[width=\linewidth]{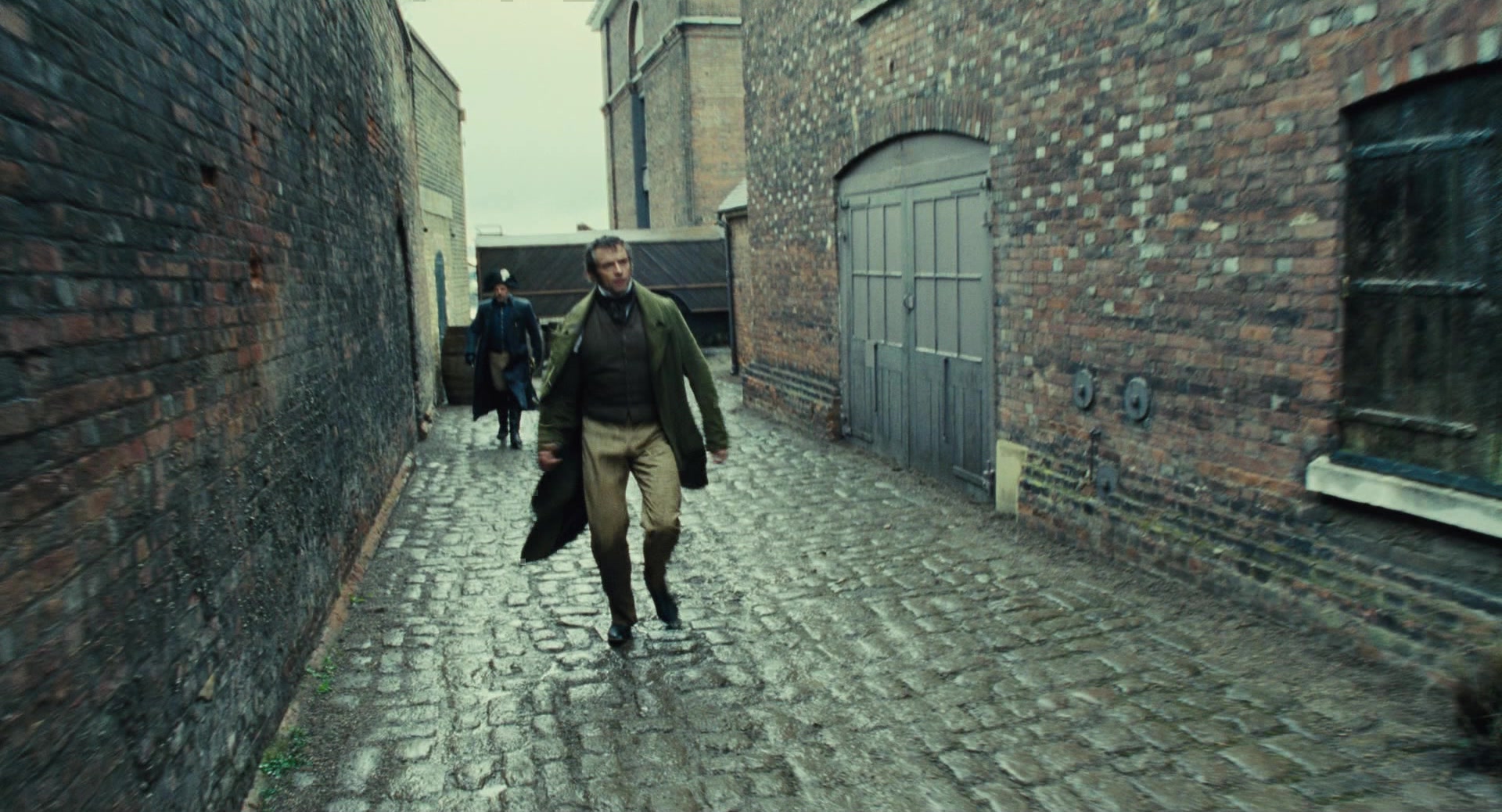} & \includegraphics[width=\linewidth]{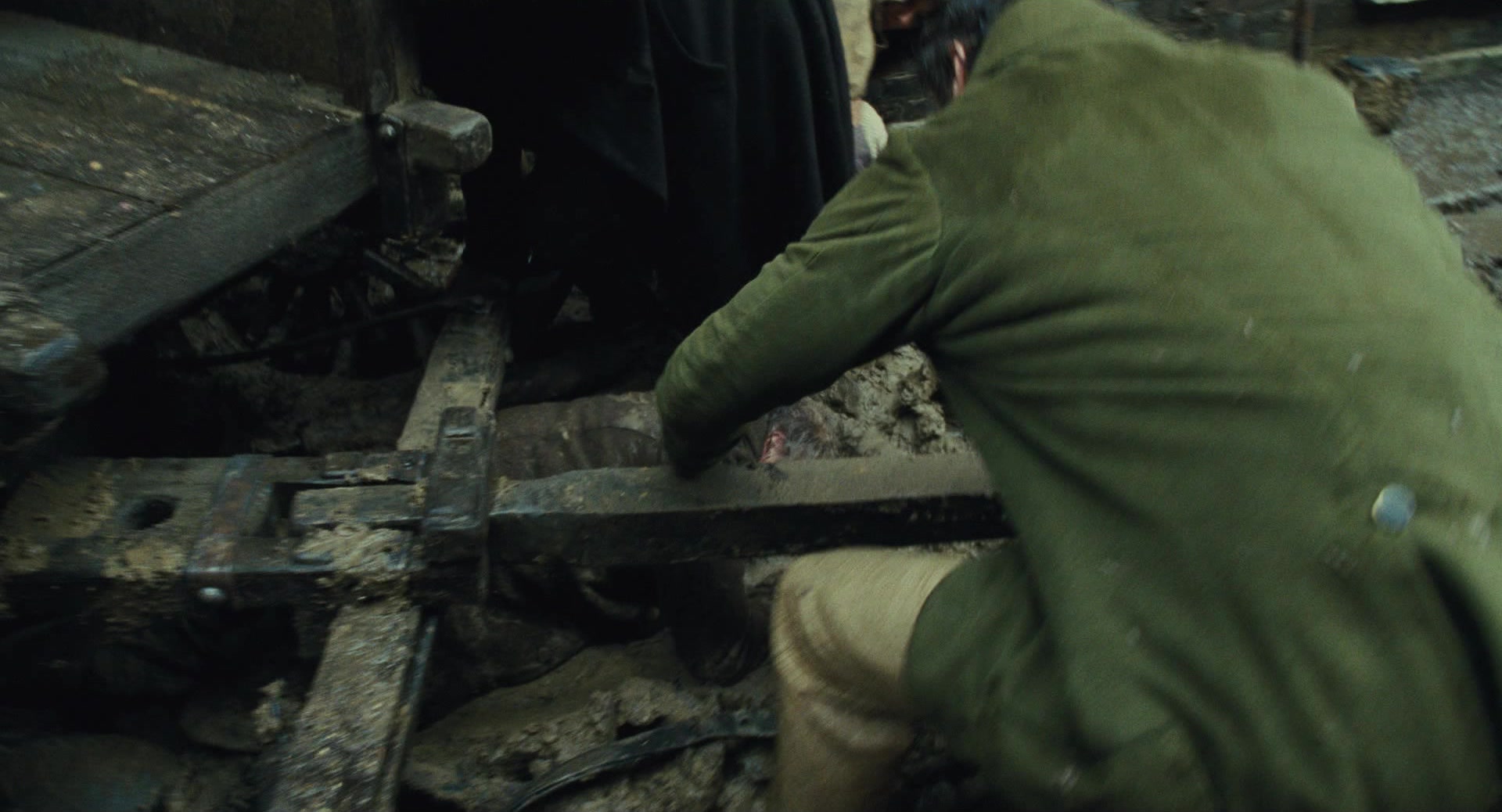} & \includegraphics[width=\linewidth]{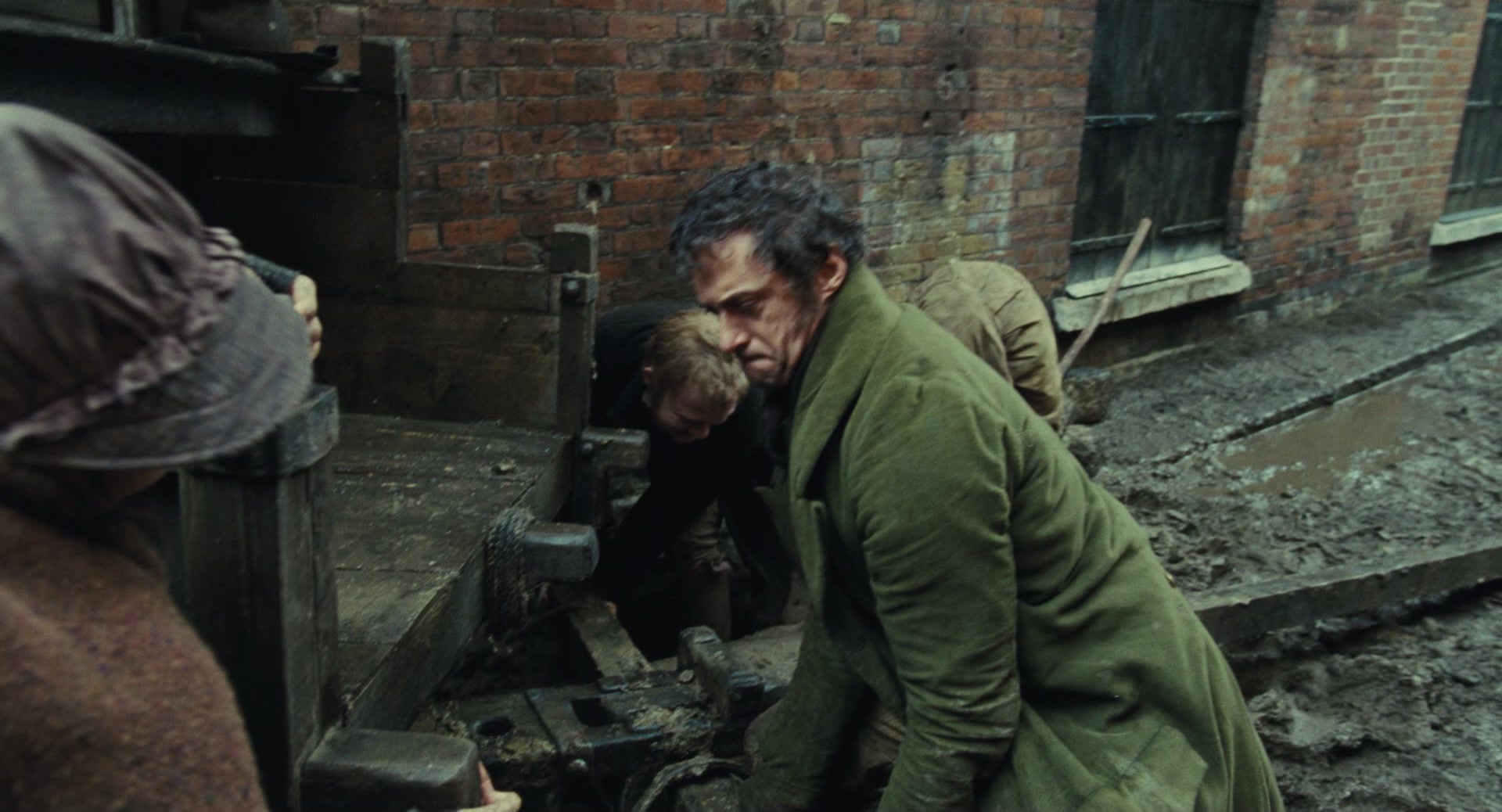} & \includegraphics[width=\linewidth]{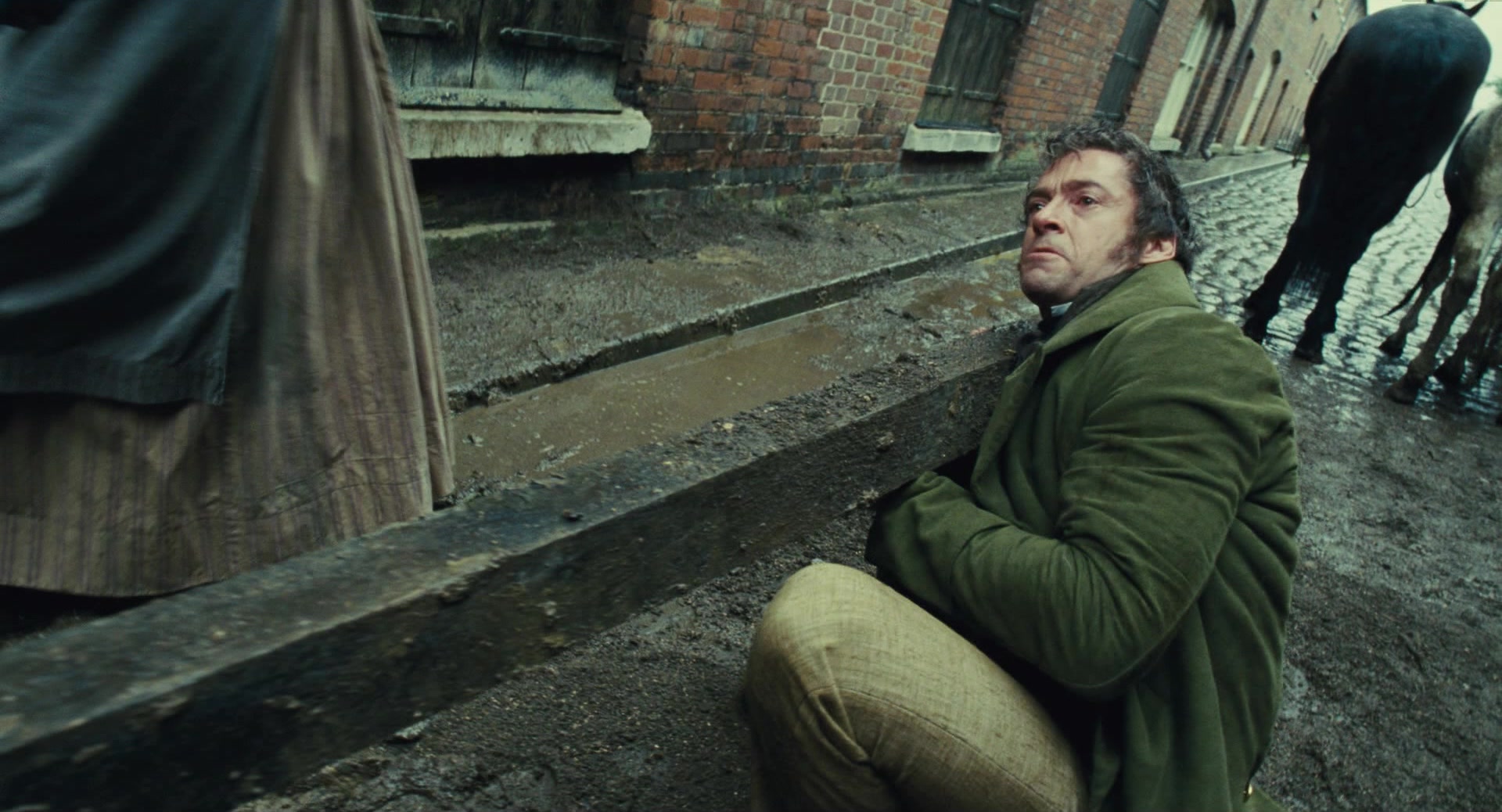} & \includegraphics[width=\linewidth]{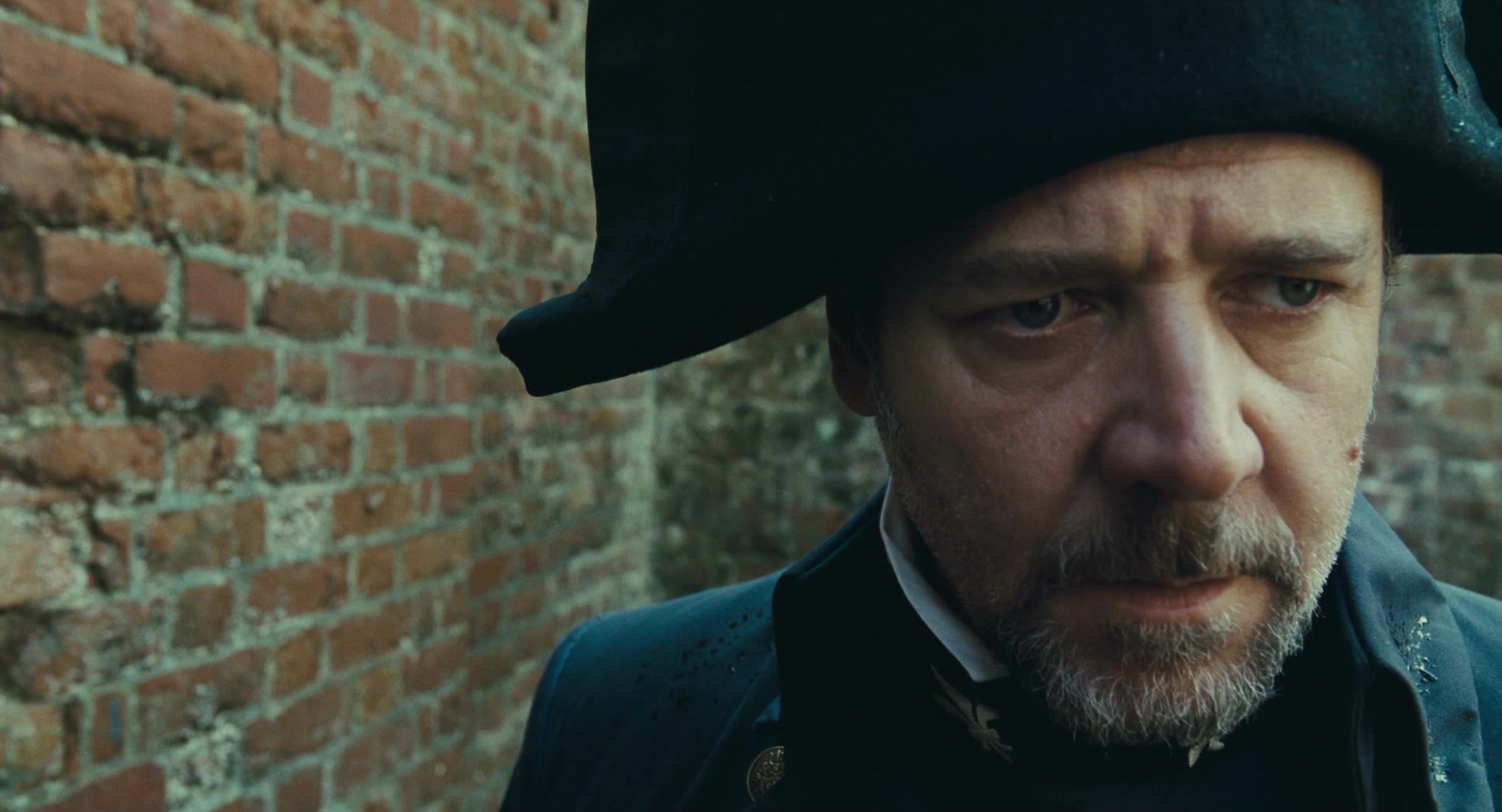}\\
 \textbf{DVS}: They rush out onto the street. & A man is trapped under a cart. &  Valjean is crouched down beside him. & Javert watches as Valjean places his shoulder under the shaft. & Javert's eyes narrow. \\
 \textbf{Script}: Valjean and Javert hurry out across the factory yard and down the muddy track beyond to discover - & A heavily laden cart has toppled onto the cart driver. &  Valjean, \redtext{Javert and Javert's assistant} all hurry to help, but they can't get a proper purchase in the spongy ground. & He throws himself under the cart at this higher end, and braces himself to lift it from beneath. & Javert stands back and looks on.\\

\end{tabular}
  \caption{Audio descriptions (DVS - descriptive video service),  movie scripts (scripts) from the movies ``Harry Potter and the prisoner of azkaban'', ``This is 40'', ``Les Miserables''. Typical mistakes contained in scripts marked with \redtext{red italic}.}
  \label{fig:teaser}
\end{center}
\end{figure*}

Figures \ref{fig:teaser1} and \ref{fig:teaser} show examples of DVS and compare them to movie scripts. Scripts have been used for various tasks \cite{laptev08cvpr,cour08eccv,marszalek09cvpr,duchenne09iccv,liang11cvpr}, but so far not for the video description. The main reason for this is that automatic alignment frequently fails due to the discrepancy between the movie and the script.
Even when perfectly aligned to the movie it frequently is not as precise as the DVS because it is typically produced prior to the shooting of the movie. \Eg. in \Figref{fig:teaser} see the mistakes marked with red. A typical case is that part of the sentence is correct, while another part contains irrelevant information.

In this work we present a novel dataset which provides transcribed DVS, which is aligned to full length HD movies. For this we retrieve audio streams from blu-ray HD disks, segment out the sections of the DVS audio and transcribe them via a crowd-sourced transcription service \cite{castingwords14}. As the audio descriptions are not fully aligned to the activities in the video, we manually align each sentence to the movie.
Therefore, in contrast to the (non public) corpus used in \cite{salway07civr, salway07corpus}, our dataset provides alignment to the actions in the video, rather than just to the audio track of the description.
In addition we also mine existing movie scripts, pre-align them automatically, similar to \cite{laptev08cvpr,cour08eccv} and then manually align the sentences to the movie.

We benchmark different approaches to generate descriptions. First are nearest neighbour retrieval using state-of-the-art visual features \cite{wang13iccv,zhou14nips,hoffman14nips} which do not require any additional labels, but retrieve sentences form the training data. Second, we propose to use semantic parsing of the sentence to extract training labels for recently proposed translation approach \cite{rohrbach13iccv} for video description.

The main contribution of this work is a novel movie description dataset which provides transcribed and aligned DVS and script data sentences. We will release sentences, alignments, video snippets, and intermediate computed features to foster research in different areas including video description, activity recognition, visual grounding, and understanding of plots.

As a first study on this dataset we benchmark several approaches for movie description. Besides sentence retrieval, we adapt the approach of \cite{rohrbach13iccv} by automatically extracting the semantic representation from the sentences using semantic parsing. This approach achieves competitive performance on TACoS Multi-Level corpus \cite{rohrbach14gcpr} without using the annotations and outperforms the retrieval approaches on our novel movie description dataset.
Additionally we present an approach to semi-automatically collect and align DVS data and analyse the differences between DVS and movie scripts.

\section{Related Work\invisible{ - 0.5 pages}}

We first discuss recent approaches to video description and then the existing works using movie scripts and DVS. 

In recent years there has been an increased interest in automatically describing images \cite{farhadi10eccv,kulkarni11cvpr,kuznetsova12acl,mitchell12eacl,li11acl,kuznetsova12acl, kuznetsova14tacl,kiros14icml,socher14tacl,fang14arxiv} and videos \cite{kojima02ijcv,gupta09cvpr,barbu12uai,hanckmann12eccvW,khan11iccvw,tan11mm,das13cvpr,guadarrama13iccv,venugopalan14arxiv,rohrbach14gcpr} with natural language. While recent works on image description show impressive results by \emph{learning} the relations between images and sentences and generating novel sentences \cite{kuznetsova14tacl,donahue14arxiv,mao14arXiv,rohrbach13iccv,kiros14arxiv,karpathy14arxiv,vinyals14arxiv,chen14arxiv}, the video description works typically rely on retrieval or templates \cite{das13cvpr,thomason14coling,guadarrama13iccv,gupta09cvpr,kojima02ijcv,kulkarni11cvpr,tan11mm} and frequently use a separate language corpus to model the linguistic statistics. A few exceptions exist: \cite{venugopalan14arxiv} uses a pre-trained model for image-description and adapts it to video description. \cite{rohrbach13iccv,donahue14arxiv} learn a translation model, however, the approaches  rely on a strongly annotated corpus with aligned videos, annotations, and sentences.
The main reason for video description lacking behind image description seems to be a missing corpus to learn and understand the problem of video description. We try to address this limitation by collecting a large, aligned corpus of video snippets and descriptions. To handle the setting of having only videos and sentences without annotations for each video snippet, we propose an approach which adapts \cite{rohrbach13iccv}, by extracting annotations from the sentences. Our extraction of annotations has similarities to \cite{thomason14coling}, but we try to extract the senses of the words automatically by using semantic parsing as discussed in \secref{sec:semantic-parsing}.

Movie 
 scripts have been used for automatic discovery and annotation of scenes and human actions in videos \cite{laptev08cvpr,marszalek09cvpr, duchenne09iccv}. We rely on the approach presented in \cite{laptev08cvpr} to align movie scripts using the subtitles.
\cite{bojanowski13iccv} attacks the problem of learning a joint model of actors and actions in movies using weak supervision provided by scripts. They also rely on a semantic parser (SEMAFOR \cite{das2012acl}) trained on FrameNet database \cite{Baker98acl}, however they limit the recognition only to two frames.
\cite{bojanowski14eccv} aims to localize individual short actions in longer clips by exploiting the ordering constrains as weak supervision.

DVS has so far mainly been studied from a linguistic prospective. \cite{salway07corpus} analyses the language properties on a non-public corpus of DVS from 91 films. Their corpus is based on the original sources to create the DVS and contains different kinds of artifacts not present in actual description, such as dialogs and production notes. In contrast our text corpus is much cleaner as it consists only of the actual DVS.
With respect to word frequency they identify that especially actions, objects, and scenes, as well as the characters are mentioned. The analysis of our corpus reveals similar statistics to theirs.

The only work we are aware of, which uses DVS in connection with computer vision is  
\cite{salway07civr}. The authors try to understand which characters interact with each other.
For this they first segment the video into events by detecting dialogue, exciting, and musical events using audio and visual features. Then they rely on the dialogue transcription and DVS to identify when characters occur together in the same event which allows them to defer interaction patterns.
In contrast to our dataset their DVS is not aligned and they try to resolve this by a heuristic to move the event which is not quantitatively evaluated. Our dataset will allow to study the quality of automatic alignment approaches, given annotated ground truth alignment.

There are some initial works to support DVS productions using scripts as source \cite{lakritz06tr} and automatically finding scene boundaries \cite{gagnon10cvprw}. However, we believe that our dataset will allow learning much more advanced multi-modal models, using recent techniques in visual recognition and natural language processing.

 Semantic parsing has received much attention in computational linguistics recently, see, for example, the tutorial \cite{Artzi:ACL2013} and references given there. Although aiming at general-purpose applicability, it has so far been successful rather for specific use-cases such as natural-language question answering \cite{Berant:EMNLP2013,Fader:KDD2014} or understanding temporal expressions \cite{Lee:ACL2014}. %

\begin{table*}[t]
\newcommand{\midruleDVSScripts}{\cmidrule(lr){1-2}  \cmidrule(lr){3-3} \cmidrule(lr){4-7}}
\center
\begin{tabular}{lrrrrrr}
\toprule
             &       & Before alignment & \multicolumn{4}{c}{After alignment} \\
             &Movies & Words            & Words           & Sentences & Avg. length & Total length\\
\midruleDVSScripts
DVS	         & 46	& 284,401 & 276,676 & 30,680 & 4.1 sec. & 34.7 h. \\
Movie script & 31	& 262,155 & 238,889 & 23,396 & 3.4 sec. & 21.7 h. \\
Total        & 72	& 546,556 & 515,565 & 54,076 & 3.8 sec. & 56.5 h. \\
\bottomrule
\end{tabular}
\vspace{-0.2cm}
\caption{Movie Description dataset statistics. Discussion see Section \ref{sec:datasetStats}.}
\vspace{-0.4cm}
\label{tab:DVS-scripts-numbers}
\end{table*}

\section{The Movie Description dataset \invisible{ - 1.5 pages}}
\label{sec:dataset}
Despite the potential benefit of DVS for computer vision, it has not been used so far apart from \cite{gagnon10cvprw, lakritz06tr} who study how to automate DVS production. We believe the main reason for this is that it is not available in the text format, \ie transcribed. We tried to get access to DVS transcripts from description services as well as movie and TV production companies, but they were not ready to provide or sell them.
While script data is easier to obtain, large parts of it do not match the movie, and they have to be ``cleaned up''. 
In the following we describe our semi-automatic approach to obtain DVS and scripts and align them to the video.

\subsection{Collection of DVS}
We search for the blu-ray movies with DVS in the ``Audio Description'' section of the British Amazon \cite{amazon14} and select a set of \nMoviesAD movies of diverse genres\footnote{\emph{2012, Bad Santa, Body Of Lies, Confessions Of A Shopaholic, Crazy Stupid Love, 27 Dresses, Flight, Gran Torino, Harry Potter and the deathly hallows Disk One, Harry Potter and the Half-Blood Prince, Harry Potter and the order of phoenix, Harry Potter and the philosophers stone, Harry Potter and the prisoner of azkaban, Horrible Bosses, How to Lose Friends and Alienate People, Identity Thief, Juno, Legion, Les Miserables, Marley and me, No Reservations, Pride And Prejudice Disk One, Pride And Prejudice Disk Two, Public Enemies, Quantum of Solace, Rambo, Seven pounds, Sherlock Holmes A Game of Shadows, Signs, Slumdog Millionaire, Spider-Man1, Spider-Man3, Super 8, The Adjustment Bureau, The Curious Case Of Benjamin Button, The Damned united, The devil wears prada, The Great Gatsby, The Help, The Queen, The Ugly Truth, This is 40, TITANIC, Unbreakable, Up In The Air, Yes man}.}.
As DVS is only available in audio format, we first retrieve audio stream from blu-ray HD disk\footnote{We use \cite{MakeMKV14} to extract a blu-ray in the .mkv file, then \cite{XMediaRecode14} to select and extract the audio streams from it.}. Then we semi-automatically segment out the sections of the DVS audio (which is mixed with the original audio stream) with the approach described below.
The audio segments are then transcribed by a crowd-sourced transcription service \cite{castingwords14} that also provides us the time-stamps for each spoken sentence. 
As the DVS is added to the original audio stream between the dialogs, there might be a small misalignment between the time of speech and the corresponding visual content. Therefore, we manually align each sentence to the movie in-house.

\begin{table*}[t]
\center
\begin{tabular}{lrrrrrr}
\toprule
Dataset& multi-sentence & domain & sentence source  & clips&videos & sentences  \\
\midrule
YouCook \cite{guadarrama13iccv} & x & cooking & crowd & &88 & 2,668 \\
TACoS \cite{regneri13tacl,rohrbach13iccv} & x & cooking & crowd &7,206&127&18,227  \\
TACoS Multi-Level \cite{rohrbach14gcpr}& x & cooking & crowd & 14,105&273 & 52,593\\
MSVD \cite{chen11acl} & & open & crowd & 1,970& & 70,028\\
Movie Description (ours) & x & open & professional & 54,076&72 & 54,076 \\ 
\bottomrule
\end{tabular}
\vspace{-0.2cm}
\caption{Comparison of video description datasets. Discussion see Section \ref{sec:datasetStats}.}
\label{tbl:datasets}
\end{table*}

\paragraph{Semi-Automatic segmentation of DVS.} 
We first estimate the temporal alignment difference between the DVS and the original audio (which is part of the DVS), as they might be off a few time frames. The precise alignment is important to compute the similarity of both streams.
Both steps (alignment and similarity) are computed using the spectograms of the audio stream, which is computed using Fast Fourier Transform (FFT).
If the difference between both audio streams is larger than a given threshold we assume the DVS contains audio description at that point in time. We smooth this decision over time using a minimum segment length of 1 second.
The threshold was picked on a few sample movies, but has to be adjusted for each movie due to different mixing of the audio description stream, different narrator voice level, and movie sound.

\subsection{Collection of script data}
\label{subsec:scripts}
In addition we mine the script web resources\footnote{http://www.weeklyscript.com, http://www.simplyscripts.com, http://www.dailyscript.com, http://www.imsdb.com} and select \nMoviesScript movie scripts\footnote{\emph{Amadeus, American Beauty, As Good As It Gets, Casablanca, Charade, Chinatown, Clerks, Double Indemnity, Fargo, Forrest Gump, Gandhi, Get Shorty, Halloween, It is a Wonderful Life, O Brother Where Art Thou, Pianist, Raising Arizona, Rear Window, The Crying Game, The Graduate, The Hustler, The Lord Of The Rings The Fellowship Of The Ring, The Lord Of The Rings The Return Of The King, The Lost Weekend, The Night of the Hunter, The Princess Bride}.}
As starting point we use the movies featuring in \cite{marszalek09cvpr} that have highest alignment scores. We are also interested in comparing the two sources (movie scripts and DVS), so we are looking for the scripts labeled as ``Final'', ``Shooting'', or ``Production Draft'' where DVS is also available. We found that the ``overlap" is quite narrow, so we analyze 5 such movies\footnote{\emph{Harry Potter and the prisoner of azkaban, Les Miserables, Signs, The Ugly Truth, This is 40}.} in our dataset. This way we end up with {31} movie scripts in total.
We follow existing approaches \cite{laptev08cvpr,cour08eccv} to automatically align scripts to movies. First we parse the scripts, extending the method of \cite{laptev08cvpr} to handle scripts which deviate from the default format. Second, we extract the subtitles from the blu-ray disks\footnote{We extract .srt from .mkv with \cite{SubtitleEdit14}. It also allows for subtitle alignment and spellchecking.}. Then we use the dynamic programming method of \cite{laptev08cvpr} to align scripts to subtitles and infer the time-stamps for the description sentences.
We select the sentences with a reliable alignment score (the ratio of matched words in the near-by monologues) of at least {0.5}. The obtained sentences are then manually aligned to video in-house.

\subsection{Statistics and comparison to other datasets}
\label{sec:datasetStats}
During the manual alignment we filter out: a) sentences describing the movie introduction/ending (production logo, cast etc); b) texts read from the screen; c) irrelevant sentences describing something not present in the video; d) sentences related to audio/sounds/music.
Table \ref{tab:DVS-scripts-numbers} presents statistics on the number of words before and after the aligment to video. One can see that for the movie scripts the reduction in number of words is about {8.9\%}, while for DVS it is {2.7\%}. In case of DVS the filtering mainly happens due to inital/ending movie intervals and transcribed dialogs (when shown as text). For the scripts it is mainly attributed to irrelevant sentences. Note, that in cases when the sentences are ``alignable'' but have minor mistakes we still keep them. 

We end up with the parallel corpus of over 50K video-sentence pairs and a total length over 56 hours. We compare our corpus to other existing parallel corpora in Table \ref{tbl:datasets}.
The main limitations of existing datasets are single domain \cite{das13cvpr,regneri13tacl,rohrbach14gcpr} or limited number of video clips \cite{guadarrama13iccv}. We fill in the gap with a large dataset featuring realistic open domain videos, which also provides high quality (professional) sentences and allows for multi-sentence description.

\subsection{Visual features}
\label{subsec:visual_features}
We extract video snippets from the full movie based on the aligned sentence intervals. We also uniformly extract 10 frames from each video snippet.
As discussed above DVS and scripts describe activities, object, and scenes (as well as emotions which we do not explicitly handle with these features, but they might still be captured, \eg by the context or activities). 
In the following we briefly introduce the visual features computed on our data which we will also make publicly available.

\textbf{DT}
We extract the improved dense trajectories compensated for camera motion \cite{wang13iccv}. For each feature (Trajectory, HOG, HOF, MBH) we create a codebook with 4000 clusters and compute the corresponding histograms. We apply L1 normalization to the obtained histograms and use them as features. 

\textbf{LSDA}
We use the recent large scale object detection CNN \cite{hoffman14nips} which distinguishes 7604 ImageNet \cite{deng09cvpr} classes. We run the detector on every second extracted frame (due to computational constraints). Within each frame we max-pool the network responses for all classes, then do mean-pooling over the frames within a video snippet and use the result as a feature.

\textbf{PLACES and HYBRID}
Finally, we use the recent scene classification CNNs \cite{zhou14nips} featuring 205 scene classes. We use both available networks: \emph{Places-CNN} and \emph{Hybrid-CNN}, where the first is trained on the Places dataset \cite{zhou14nips} only, while the second is additionally trained on the 1.2 million images of ImageNet (ILSVRC 2012) \cite{ILSVRCarxiv14}. We run the classifiers on all the extracted frames of our dataset. 
We mean-pool over the frames of each video snippet, using the result as a feature.

\section{Approaches to video description \invisible{ - 0.5 page}}
\label{sec:approaches}
In this section we describe the approaches to video description that we benchmark on our proposed dataset.

\textbf{Nearest neighbor}
We retrieve the closest sentence from the training corpus using the L1-normalized visual features introduced in Section \ref{subsec:visual_features} and the intersection distance.  

\textbf{SMT}
We adapt the two-step translation approach of \cite{rohrbach13iccv} which uses an intermediate semantic representation (SR), modeled as a tuple, \eg $\langle cut,knive,tomato \rangle$.  As the first step it learns a mapping from the visual input to the semantic representation (SR), modeling pairwise dependencies in a CRF using visual classifiers as unaries. The unaries are trained using an SVM on dense trajectories \cite{wang13ijcv}.  
In the second step \cite{rohrbach13iccv} translates the SR to a sentence using  Statistical Machine Translation (SMT)~\cite{koehn07acl}. For this the approach concatenates SR as input language, \eg \emph{cut knife tomato}, and the natural sentence pairs as output language, \eg \emph{The person slices the tomato.} While we cannot rely on an annotated SR as in \cite{rohrbach13iccv}, we automatically mine the SR from sentences using semantic parsing which we introduce in the next section.
 In addition to dense trajectories we use the features described in \secref{subsec:visual_features}.

\textbf{SMT Visual words}
As an alternative on potentially noisy labels extracted from the sentences, we try to directly translate visual classifiers and visual words to a sentence.  
We model the essential components by relying on activity, object, and scene recognition. For objects and scenes we rely on the pre-trained models LSDA and PLACES. %
For activities we rely on the state-of-the-art activity recognition feature DT. We cluster the DT histograms to 300 visual words using k-means. The index of the closest cluster center from our activity category is chosen as label.
To build our tuple we obtain the highest scoring  class labels
of the object detector and scene classifier. More specifically for the object detector we consider two highest scoring classes: for subject and object. 
Thus we obtain the tuple $\langle SUBJECT, ACTIVITY, OBJECT, SCENE \rangle = \langle argmax(LSDA), DT_{i},argmax2(LSDA),$ $ argmax(PLACES)\rangle $, for which we learn translation to a natural sentence using the SMT approach discussed above.

\newcommand{\q}[1] {``\textit{#1}''}
\newcommand{\qp}[1] {\textit{#1}}
\newcommand{\lbl}[1] {\texttt{\small #1}}
\newcommand{\ignore}[1] {}

\section{Semantic parsing}
\label{sec:semantic-parsing}

\begin{table}
\center
\small
\begin{tabular}{p{1.8cm} p{1cm} p{1.7cm} p{2.1cm}}
\toprule
Phrase        &       WordNet            & VerbNet                  & Expected \\
              &       Mapping            & Mapping                  & Frame \\
\midrule
the man       &         man\#1           & Agent.animate            & Agent: man\#1\\
\cmidrule(lr){1-4}
begin to shoot&         shoot\#2         & shoot\#vn\#2             & Action: shoot\#2\\
\cmidrule(lr){1-4}
a video       &         video\#1         & Patient.solid            & Patient: video\#1\\
\cmidrule(lr){1-4}
in            &         in               & PP.in                    & \\
\cmidrule(lr){1-4}
the moving bus&         bus\#1           & NP.Location. solid        & Location: moving bus\#1\\
\bottomrule
\end{tabular}
\caption{Semantic parse for \q{He began to shoot a video in the moving bus}. Discussion see Section \ref{sec:semparsingAppraoch}}
\label{tab:semantic-parse-expected-output}
\end{table}

Learning from a parallel corpus of videos and sentences without having annotations is challenging. 
In this section we introduce our approach to exploit the sentences using semantic parsing. The proposed method aims to extract annotations from the natural sentences and make it possible to avoid the tedious annotation task.
Later in the section we perform the evaluation of our method on a corpus where annotations are available in context of a video description task.

\subsection{Semantic parsing approach}
\label{sec:semparsingAppraoch}

We lift the words in a sentence to a semantic space of roles and WordNet \cite{pedersen2004wordnet,Fellbaum1998} senses by performing SRL (Semantic Role Labeling) and WSD (Word Sense Disambiguation). For an example, refer to Table \ref{tab:semantic-parse-expected-output}, the expected outcome of semantic parsing on the input sentence \q{He shot a video in the moving bus} is ``\lbl{Agent: man, Action: shoot, Patient: video, Location: bus}''. Additionally, the role fillers are disambiguated.

We use the ClausIE tool \cite{clauseIE} to decompose sentences into their respective clauses. For example, \q{he shot and modified the video} is split into two phrases \q{he shot the video} and \q{the modified the video}). We then use the OpenNLP tool suite\footnote{http://opennlp.sourceforge.net/} for chunking
the text of each clause. In order to provide the linking of words in the sentence to their WordNet sense mappings, we rely on a state-of-the-art WSD system, IMS \cite{ims-wsd}. The WSD system, however, works at a word level. We enable it to work at a phrase level. For every noun phrase, we identify and disambiguate its head word (\eg \lbl{the moving bus} to \q{bus\#1}, where \q{bus\#1} refers to the first sense of the word \lbl{bus}). We link verb phrases to the proper sense of its head word in WordNet (\eg \lbl{begin to shoot} to \q{shoot\#2}).

In order to obtain word role labels, we link verbs to VerbNet \cite{verbnet-2009,verbnet-2006}, a manually curated high-quality linguistic resource for English verbs. VerbNet is already mapped to WordNet, thus we map to VerbNet via WordNet. We perform two levels of matches in order to obtain role labels. First is the syntactic match. Every VerbNet verb sense comes with a syntactic frame \eg for \lbl{shoot}, the syntactic frame is \lbl{NP V NP}. We first match the sentence's verb against the VerbNet frames. These become candidates for the next step.
Second we perform the semantic match: VerbNet also provides a role restriction on the arguments of the roles \eg for \lbl{shoot} (sense killing), the role restriction is \lbl{Agent.animate V Patient.\textbf{animate} PP Instrument.solid}. The other sense for \lbl{shoot} (sense snap), the semantic restriction is \lbl{Agent.animate V Patient.\textbf{solid}}. We only accept candidates from the syntactic match that satisfy the semantic restriction.

\begin{figure}[t]
\begin{center}
\begin{subfigure}[b]{\linewidth}
  \includegraphics[width=0.9\linewidth]{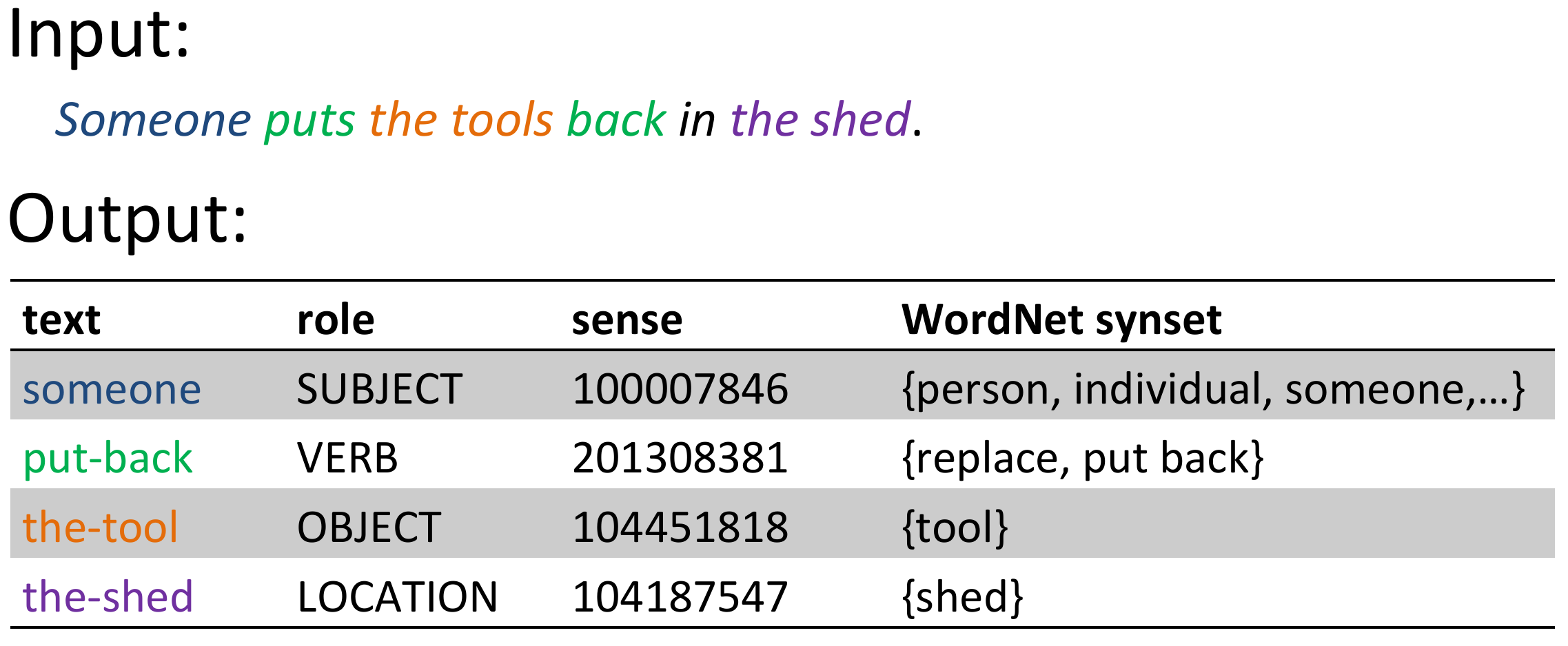}
  \caption[labelInTOC]{Semantic representation extracted from a sentence.}
  \label{fig:semp1}
  \end{subfigure}
\begin{subfigure}[b]{\linewidth}
  \includegraphics[width=0.8\linewidth]{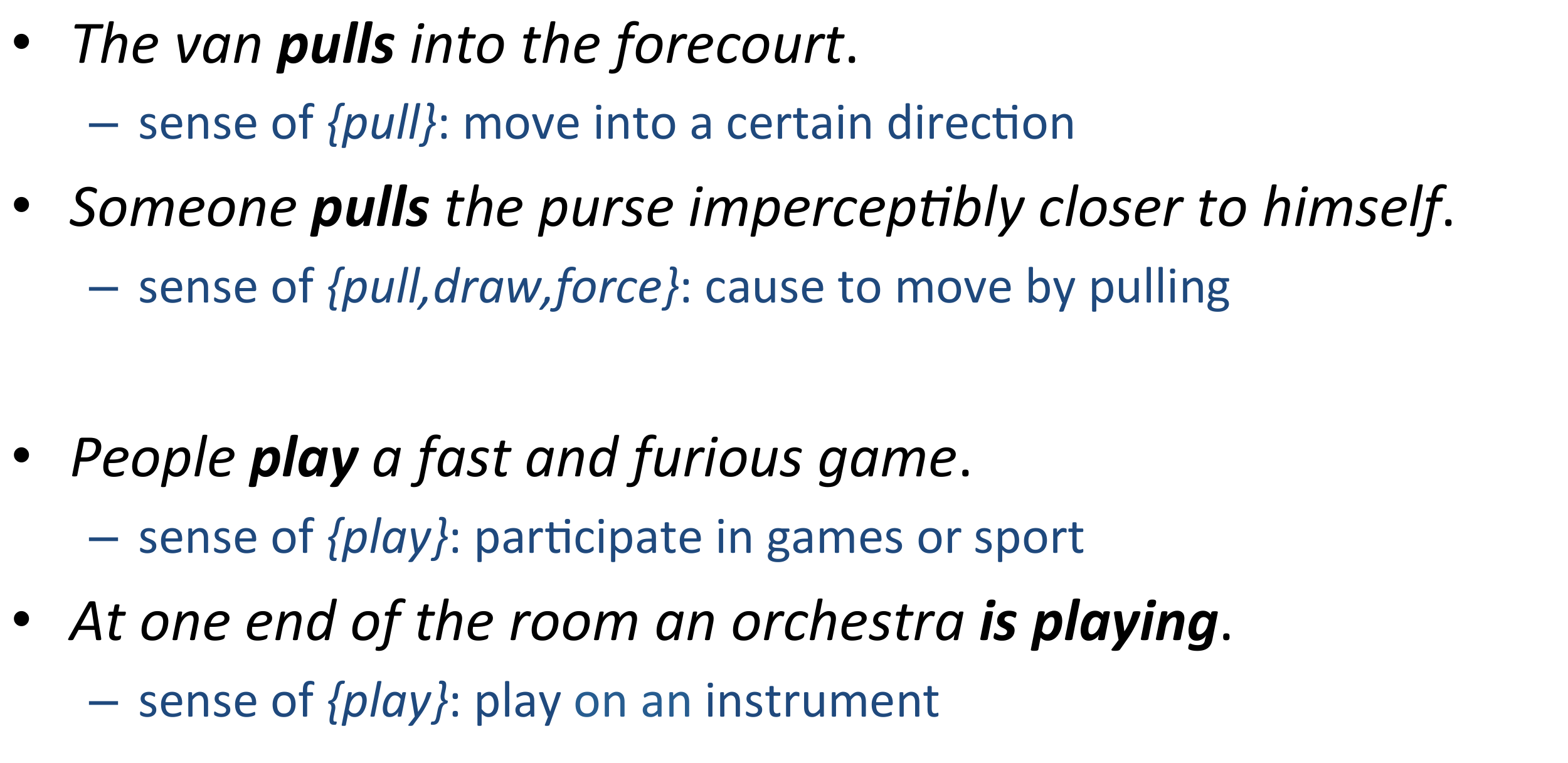}
  \caption[labelInTOC]{Same verb, different senses.}
  \label{fig:semp2}
 \end{subfigure}
\begin{subfigure}[b]{\linewidth}
  \includegraphics[width=0.8\linewidth]{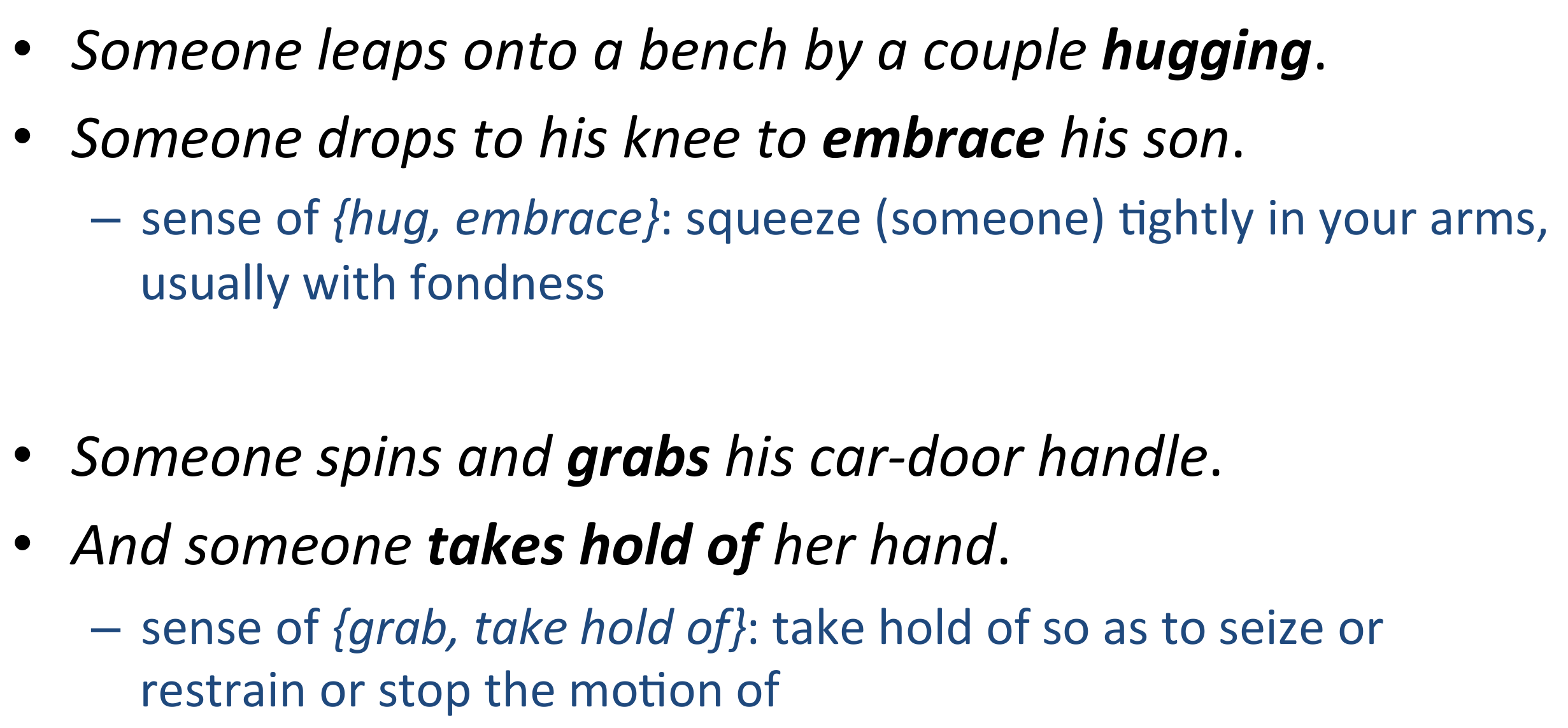}
  \caption[labelInTOC]{Different verbs, same sense.}
  \label{fig:semp3}
    \end{subfigure}
    \caption[labelInTOC]{Semantic parsing example, see Section \ref{sec:semparsingAppraoch}}
\end{center}
\end{figure}

VerbNet contains over 20 roles and not all of them are general or can be recognized reliably. Therefore, we further group them to get the SUBJECT, VERB, OBJECT and LOCATION roles.
We explore two approaches to obtaining the labels based on the output of the semantic parser. First is to use the extracted text chunks directly as labels. Second is to use the corresponding senses as a labels (and therefore group multiple text labels). In the following we refer to these as \emph{text-} and \emph{sense-labels}.
Thus from each sentence we extract a semantic representation in a form of (SUBJECT, VERB, OBJECT, LOCATION), %
see Figure~\ref{fig:semp1} for example. 
Using the WSD allows to identify different senses (WordNet synsets) for the same verb
(Figure~\ref{fig:semp2}) and the same sense for different verbs (Figure~\ref{fig:semp3}).

\subsection{Applying parsing to TACoS Multi-Level corpus}
\label{sec:applyTACOS}

\newcommand{\midruleResTrans}{\cmidrule(lr){1-1}\cmidrule(lr){2-2}} %
\begin{table}[t]
\center
\begin{tabular}{lrr}
\toprule
Approach              & BLEU \\
\midruleResTrans
SMT \cite{rohrbach13iccv} & 24.9 \\
SMT \cite{rohrbach14gcpr} & 26.9 \\
SMT with our text-labels     & 22.3 \\
SMT with our sense-labels    & 24.0 \\
\bottomrule
\end{tabular}
\caption{BLEU@4 in \% on sentences of Detailed~Descriptions of the TACoS Multi-Level \cite{rohrbach14gcpr} corpus, see Section \ref{sec:applyTACOS}.}
\label{tbl:res:detailed}
\figvspace
\end{table}

We apply the proposed semantic parsing to the TACoS Multi-Level \cite{rohrbach14gcpr} parallel corpus. We extract the SR from the sentences as described above and use those as annotations. Note, that this corpus is annotated with the tuples (ACTIVITY, OBJECT, TOOL, SOURCE, TARGET) and the subject is always the person. Therefore we drop the SUBJECT role and only use (VERB, OBJECT, LOCATION) as our SR.
Then, similar to \cite{rohrbach14gcpr}, we train the visual classifiers for our labels (proposed by the parser), we only use the ones that appear at least 30 times. Next we train a CRF with 3 nodes for verbs, objects and locations, using the visual classifier responses as unaries. We follow the translation approach of \cite{rohrbach13iccv} and train the SMT on the Detailed Descriptions part of the corpus using our labels. Finally, we translate the SR predicted by our CRF to generate the sentences.
Table \ref{tbl:res:detailed} shows the results comparing our method to \cite{rohrbach13iccv} and \cite{rohrbach14gcpr} who use manual annotations to train their models. As we can see the sense-labels perform better than the text-labels as they provide better grouping of the labels. Our method produces competitive result which is only {0.9\%} below the result of \cite{rohrbach13iccv}. At the same time \cite{rohrbach14gcpr} uses more training data, additional color Sift features and recognizes the dish prepared in the video. All these points, if added to our approach, would also improve the performance.

\newcommand{\midrulestat}{\cmidrule(lr){1-1}\cmidrule(lr){2-6}}%

\begin{table}[t]
\center
\begin{tabular}{p{2.5cm} p{0.8cm} p{0.3cm} p{0.7cm} p{0.7cm} p{0.7cm} }
\toprule

Annotations & {activity} & {tool}  & {object} & {source} & {target} \\
\midrulestat
Manual \cite{rohrbach14gcpr} & 78 & 53 & 138 & 69 & 49 \\
\midrule
 & {verb} & \multicolumn{2}{c}{object} & \multicolumn{2}{c}{location} \\
\cmidrule(lr){2-6}
Our text-labels                   & 145 & \multicolumn{2}{c}{260} & \multicolumn{2}{c}{85} \\
Our sense-labels                  & 158 & \multicolumn{2}{c}{215} & \multicolumn{2}{c}{85} \\
\bottomrule 
\end{tabular}
\caption{Label statistics from our semantic parser on TACoS Multi-Level \cite{rohrbach14gcpr} corpus, see Section \ref{sec:applyTACOS}.}
\label{tbl:crfnodes}
\figvspace
\end{table}

We analyze the labels selected by our method in Table \ref{tbl:crfnodes}. It is clear that our labels are still imperfect, \ie different labels might be assigned to similar concepts. However the number of extracted labels is quite close to the number of manual labels. Note, that the annotations were created prior to the sentence collection, so some verbs used by humans in sentences might not be present in the annotations.

From this experiment we conclude that the output of our automatic parsing approach can serve as a replacement of manual annotations and allows to achieve competitive results. In the following we apply this approach to our movie description dataset.

\section{Evaluation \invisible{ - 1.2 pages}}
In this section we provide more insights about our movie description dataset. First we compare DVS to movie script and then we benchmark the approaches to video description introduced in Section \ref{sec:approaches}.

\subsection{Comparison DVS vs script data}
\label{sec:comparisionDVS}
We compare the DVS and script data using {5} movies from our dataset where both are available (see Section \ref{subsec:scripts}).   
For these movies we select the overlapping time intervals with the intersection over union overlap of at least {75\%}, which results in {126} sentence pairs. We ask humans via Amazon Mechanical Turk (AMT) to compare the sentences with respect to their correctness and relevance to the video, using both video intervals as a reference (one at a time, resulting in 252 tasks). Each task was completed by 3 different human subjects.
Table \ref{tab:DVS-scripts} presents the results of this evaluation. DVS is ranked as more correct and relevant in over {60\%} of the cases, which supports our intuition that scrips contain mistakes and irrelevant content even after being cleaned up and manually aligned.

\begin{table}[t]
\center
\begin{tabular}{lll}
\toprule
                & Correctness & Relevance \\
\midrule
DVS	            & 63.0	     & 60.7	\\
Movie scripts	& 37.0	     & 39.3	\\
\bottomrule
\end{tabular}
\caption{Human evaluation of DVS and movie scripts: which sentence is more correct/relevant with respect to the video, in \%. Discussion in Section \ref{sec:comparisionDVS}.}
\label{tab:DVS-scripts}
\end{table}

\begin{table}
\centering
\begin{tabular}{@{\ }ll@{\ \ }l@{\ \ \ }l@{\ \ }l@{\ }}
\toprule
Corpus &Clause&NLP&Labels&WSD \\
\midrule
TACoS Multi-Level \cite{rohrbach14gcpr} & 0.96  & 0.86  & 0.91  & 0.75  \\
Movie Description (ours) & 0.89  & 0.62  & 0.86  & 0.7  \\
\bottomrule
\end{tabular}
\caption{Semantic parser accuracy for TACoS Multi-Level and our new corpus. Discussion in Section \ref{sec:semanticParserEval}.}
\label{tab:semantic-parse-accuracy-per-source-detailed}
\end{table}

\subsection{Semantic parser evaluation}
\label{sec:semanticParserEval}
Table \ref{tab:semantic-parse-accuracy-per-source-detailed} reports the accuracy of the different components of the semantic parsing pipeline. The components are clause splitting (Clause), POS tagging and chunking (NLP), semantic role labeling (Labels) and word sense disambiguation (WSD). We manually evaluate the correctness on a randomly sampled set of sentences using human judges. %
It is evident that the poorest performing parts are the NLP and the WSD components. Some of the NLP mistakes arise due to incorrect POS tagging. WSD is considered a hard problem and when the dataset contains less frequent words, the performance is severely affected. Overall we see that the movie description corpus is more challanging than TACoS Multi-Level but the drop in performance is reasonable compared to the siginificantly larger variability.

\begin{table}[t]
\center
\begin{tabular}{p{2.8cm} p{1.5cm} p{1.2cm} p{1.3cm}}
\toprule
                & Correctness & Grammar & Relevance \\
\midrule
Nearest neighbor\\
DT			    &7.6	&5.1	&7.5	\\
LSDA		&7.2	&4.9	&7.0	\\
PLACES	&7.0	&5.0	&7.1	\\
HYBRID	&6.8	&4.6	&7.1	\\
\midrule
SMT Visual words:	&7.6	&8.1	&7.5	\\
\midrule
\multicolumn{4}{l}{SMT with our text-labels}\\
DT 30	&6.9	&8.1	&6.7	\\
DT 100	&5.8	&6.8	&5.5	\\
All 100	&4.6	&5.0	&4.9	\\
\midrule
\multicolumn{4}{l}{SMT with our sense-labels}\\
DT 30	&6.3	&6.3	&5.8	\\
DT 100	&4.9	&5.7	&5.1	\\
All 100 &5.5	&5.7	&5.5	\\
\midrule
Movie script/DVS	&2.9	&4.2	&3.2	\\
\bottomrule
\end{tabular}
\caption{Comparison of approaches. Mean Ranking (1-12). Lower is better. Discussion in Section \ref{sec:VideoDescription}.}
\label{tab:humaneval}
\end{table}

\subsection{Video description}
\label{sec:VideoDescription}

As the collected text data comes from the movie context, it contains a lot of information specific to the plot, such as names of the characters. We pre-process each sentence in the corpus, transforming the names and other person related information (such as ``a young woman'') to ``someone'' or ``people''. The transformed version of the corpus is used in all the experiments below. We will release the transformed and the original corpus.

We use the {5} movies mentioned before (see Section \ref{subsec:scripts}) as a test set for the video description task, while all the others (67) are used for training.
Human judges were asked to rank multiple sentence outputs with respect to their correctness, grammar and relevance to the video. 

Table \ref{tab:humaneval} summarizes results of the human evaluation from 250 randomly selected test video snippets, showing the mean rank, where lower is better. 
In the top part of the table we show the nearest neighbor results based on multiple visual features. When comparing the different features, we notice that the pre-trained features (LSDA, PLACES, HYBRID) perform better than DT, where HYBRID performing best. Next is the translation approach with the visual words as labels, performing overall worst of all approaches. The next two blocks correspond to the translation approach when using the labels from our semantic parser. After extracting the labels we select the ones which appear at least 30 or 100 times as our visual attributes. As 30 results in a much higher number of attributes (see Table~\ref{tbl:crfnodes_movies})  predicting the SR turns into a more difficult recognition task, resulting in worse mean rankings. ``All 100'' refers to combining all the visual features as unaries in the CRF. Finally, the last ``Movie script/DVS'' block refers to the actual test sentences from the corpus and not surprisingly ranks best.

\begin{figure*}[t]
\begin{center}
\begin{tabular}{cccc}
 \includegraphics[width=4cm]{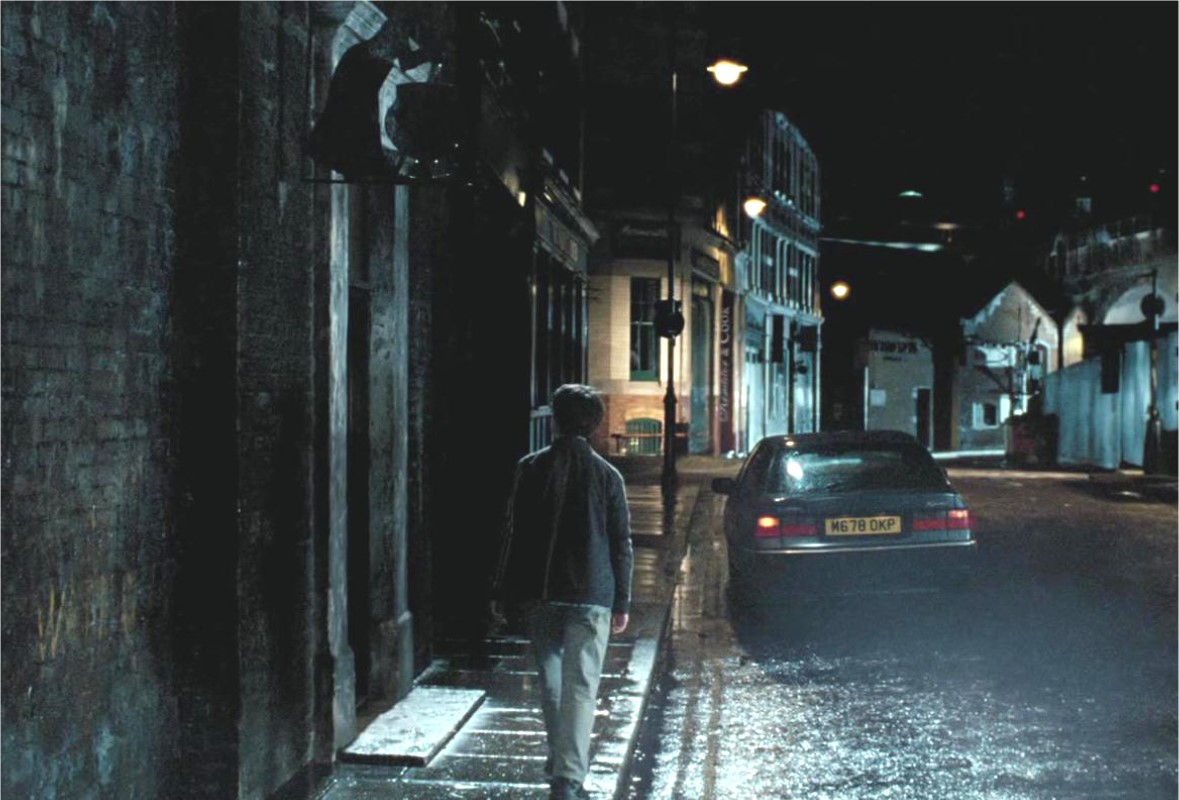} &
 \includegraphics[width=4cm]{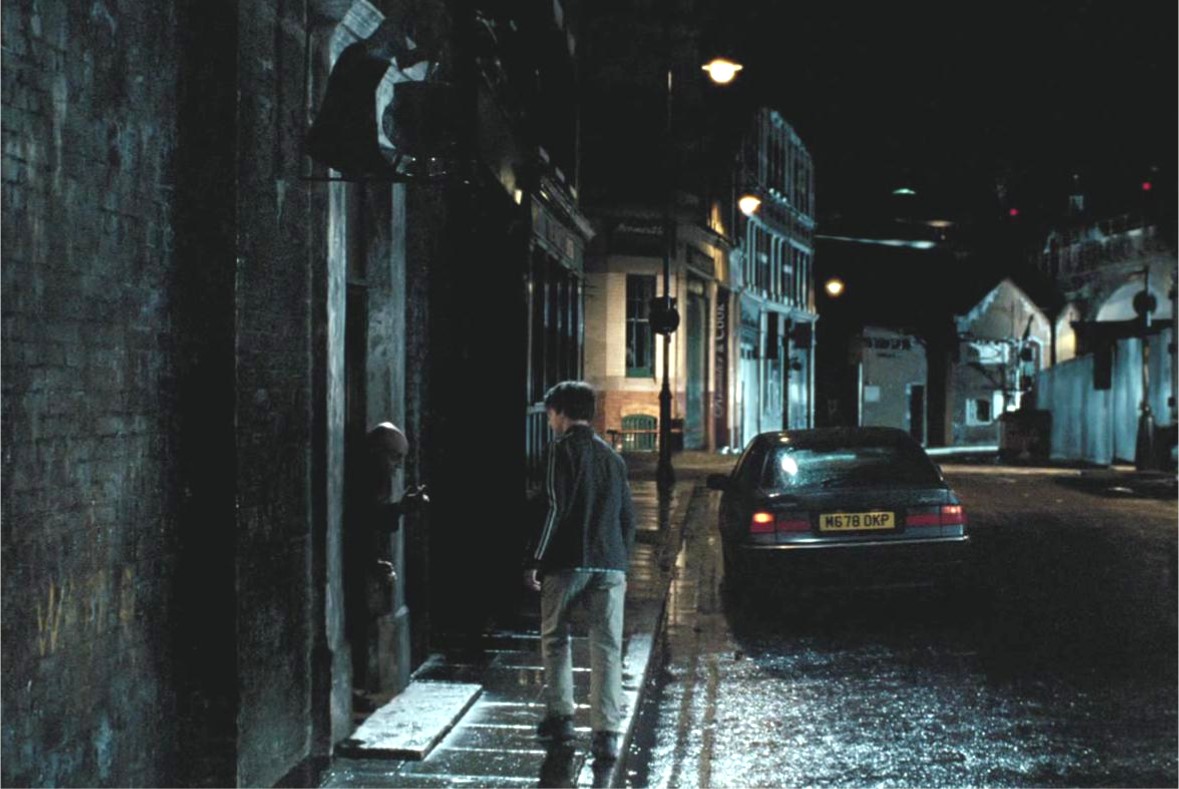} &
 \includegraphics[width=4cm]{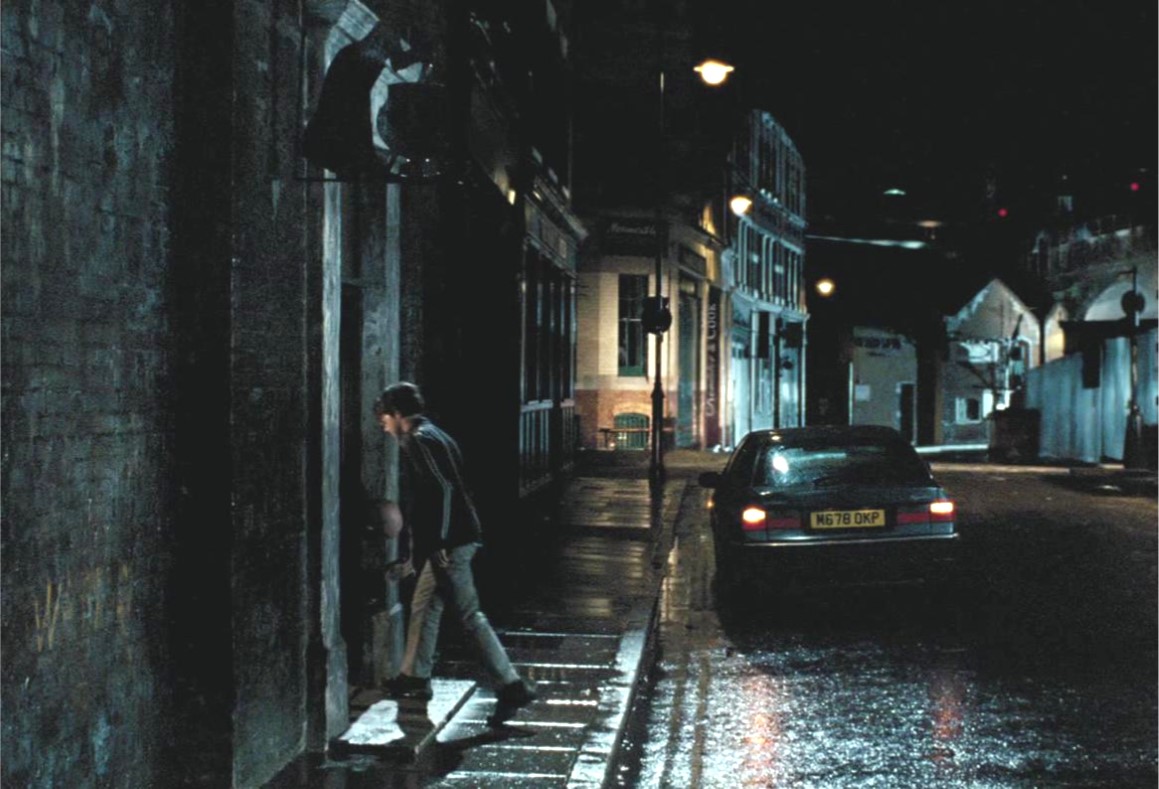} &
 \includegraphics[width=4cm]{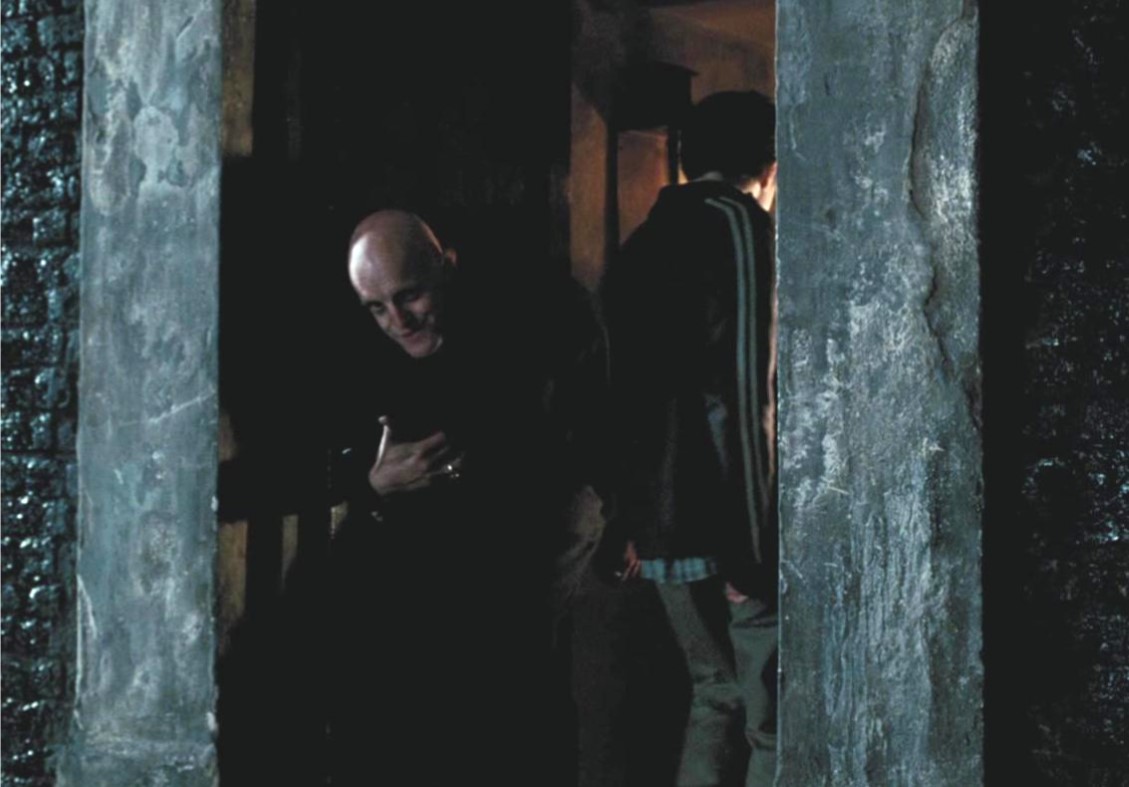}\\
 \end{tabular}
 \begin{tabular}{ll}
 Nearest neighbor\\
 DT			& People stand with a happy group, including someone.\\
 LSDA		&    The hovering Dementors chase the group into the lift.\\
 HYBRID	&    Close by, a burly fair-haired someone in an orange jumpsuit runs down a dark street.\\
 PLACES	&    Someone is on his way to look down the passage way between the houses.\\
\midrule
SMT Visual words &   Someone in the middle of the car pulls up ahead\\
\midrule
\multicolumn{2}{l}{SMT with our text-labels}\\
 DT 30	&    Someone opens the door to someone\\
 DT 100	&    Someone, the someone, and someone enters the room \\
 All 100	&        Someone opens the door and shuts the door, someone and his someone \\
\midrule
\multicolumn{2}{l}{SMT with our sense-labels}\\
 DT 30	&   Someone, the someone, and someone enters the room \\
 DT 100	&    Someone goes over to the door\\
 All 100	&       Someone enters the room\\
\midrule
Movie script/DVS	&    Someone follows someone into the leaky cauldron \\
 \end{tabular}
 \end{center}
  \caption{Qualitative comparison of different video description methods. Discussion in Section \ref{sec:VideoDescription}. More examples on our web page.}
  \label{fig:qual}
\end{figure*}

Overall we can observe three main tendencies: (1) Using our parsing with SMT outperforms nearest neighbor baselines and SMT Visual words. (2) In contrast to the kitchen dataset, the sense labels perform slightly worse than the text labels, which we attribute to the errors made in the WSD. (3) The actual movie script/DVS are ranked on average significantly better than any of the automatic approaches. These tendencies are also reflected in Figure \ref{fig:qual}, showing example outputs of all the evaluated approaches for a single movie snippet. Examining more qualitative examples which we provide on our web page indicates that it is possible to learn relevant information from this corpus.

\begin{table}[t]
\center
\begin{tabular}{p{2.3cm} p{1cm} p{1cm} p{1cm} p{1cm} }
\toprule
Annotations     & {subject} & {verb} & {object} & {location} \\
\midrule
text-labels 30   & 24 & 380 & 137 & 71 \\
sense-labels 30  & 47 & 440 & 244 & 110 \\
text-labels 100  & 8  & 121 & 26 & 8 \\
sense-labels 100 & 8  & 143 & 51 & 37 \\
\bottomrule 
\end{tabular}
\caption{Label statistics from our semantic parser on the movie description corpus. 30 and 100 indicate the minimum number of label occurrences in the corpus, see Section \ref{sec:VideoDescription}.}
\label{tbl:crfnodes_movies}
\figvspace
\end{table}

\section{Conclusions \invisible{ - 0.3 pages}}

In this work we presented a novel dataset of movies with aligned descriptions sourced from movie scripts and DVS (audio descriptions for the blind).
We present first experiments on this dataset using state-of-the art visual features, combined with a recent movie description approach from \cite{rohrbach13iccv}. We adapt the approach for this dataset to work without annotations, but rely on semantic parsing of labels. We show competitive performance on the TACoS Multi-Level dataset and promising results on our movie description data.
We compare DVS with previously used script data and find that DVS tends to be more correct and relevant to the movie than script sentences.
Beyond our first study on single sentences, the dataset opens new possibilities to understand stories and plots across multiple sentences in an open domain scenario on large scale. Something no other video nor image description dataset can offer as of now.

\section{Acknowledgements}
Marcus Rohrbach was supported by a fellowship within the FITweltweit-Program of the German Academic Exchange Service (DAAD).

\small
\bibliographystyle{ieee}
\bibliography{biblioLong,rohrbach,rohrbach15cvpr}

\end{document}